\documentclass[11pt]{article}

\usepackage[para,online,flushleft]{threeparttable}
\usepackage{soul}
\usepackage{sidecap}
\usepackage{epsfig}
\usepackage{color}
\usepackage{graphicx}
\usepackage{multirow}
\usepackage{hanging}
\usepackage{wrapfig}
\usepackage{subfig}
\usepackage{epstopdf}
\usepackage{float}
\usepackage{booktabs}
\usepackage[small,compact]{titlesec}
\usepackage{palatino}
\usepackage{url}
\usepackage[rflt]{floatflt}
\usepackage{subfig}
\usepackage{comment}
\usepackage{amssymb}
\usepackage{multirow}
\usepackage{cite}
\usepackage{tabularx}
\usepackage{xcolor}
\usepackage{listings}
\usepackage{mscp}
\graphicspath{{./figures/} }
\usepackage{multicol}
\usepackage{threeparttable}

\newcommand{\blue}[1]{#1}

\definecolor{dkgreen}{rgb}{0,0.6,0}
\definecolor{gray}{rgb}{0.5,0.5,0.5}
\definecolor{mauve}{rgb}{0.58,0,0.82}

\newcounter{myctr}

\newcommand{\ignore}[1]{}

\usepackage{xspace}
\newcommand{\projectname}{CoCo-Gen\xspace}
\begin{document}
\title{CoCoPIE: Making Mobile AI Sweet As PIE\\
---Compression-Compilation Co-Design Goes a Long Way}

\author{Shaoshan Liu${}^\dagger$, Bin Ren${}^\ast$, Xipeng Shen${}^\ddagger$, Yanzhi Wang${}^\diamond$\\
${}^\dagger$Perceptin Inc.\hspace{.1in} ${}^\ast$William \& Mary\hspace{.1in}\\ ${}^\ddagger$North Carolina State University\hspace{.1in} ${}^\diamond$Northeastern University\\
Contact: info@cocopie.ai
}

\date{}

\maketitle
\begin{abstract}
    Assuming hardware is the major constraint for enabling real-time mobile intelligence, the industry has mainly dedicated their efforts to developing specialized hardware accelerators for machine learning and inference. This article challenges the assumption. By drawing on a recent real-time AI optimization framework CoCoPIE, it maintains that with effective compression-compiler co-design, it is possible to enable real-time artificial intelligence on mainstream end devices without special hardware. CoCoPIE is a software framework that holds numerous records on mobile AI: the first framework that supports all main kinds of DNNs, from CNNs to RNNs, transformer, language models, and so on; the fastest DNN pruning and acceleration framework, up to 180X faster compared with current DNN pruning on other frameworks such as TensorFlow-Lite; making many representative AI applications able to run in real-time on off-the-shelf mobile devices that have been previously regarded possible only with special hardware support; making off-the-shelf mobile devices outperform a number of representative ASIC and FPGA solutions in terms of energy efficiency and/or performance.
\end{abstract}

It has been a consensus that the company who enables real intelligence on end devices (such as mobile devices and IoT devices) will define the future of computing. Racing towards this goal, many companies, whether giant technology firms such as Google, Microsoft, Amazon, Apple and Facebook, or startups spent tens of billions of dollars each year on R\&D. 
Assuming hardware is the major constraint for enabling real-time mobile intelligence, the industry has mainly dedicated their efforts to developing specialized hardware accelerators for machine learning and inference. Billions of dollars have been spent to fuel this intelligent hardware race. 

This article challenges the assumption. By drawing on a recent real-time AI optimization framework CoCoPIE, it maintains that with effective {\em compression-compiler co-design}, it is possible to enable real-time artificial intelligence (AI) on mainstream end devices without special hardware. 

The principle of {\em compression-compilation co-design} is to design the compression of Deep Learning Models and their compilation to executables in a hand-in-hand manner. This synergistic method can effectively optimize both the size and speed of Deep Learning models, and also can dramatically shorten the tuning time of the compression process, largely reducing the time to the market of AI products. When applied to models running on mainstream end devices, the method can produce real-time experience across a set of AI applications that had been broadly perceived possible only with special AI accelerators.






Foregoing the need for special hardware for real-time AI has some profound implications, thanks to the multi-fold advantages of mainstream processors over special hardware:  

\begin{itemize}
\item Time to market: Special hardware often takes multiple years before it reaches the market. The creation of the associated compiler and system software further lengthens the process. Applications using such hardware often needs to use the special APIs and meet many special constraints (e.g., tiling computations to a certain size), which lengthens the time to market of AI product. 

\item Cost: Developing a special ASIC processor is costly, and adding them into existing systems incurs extra expenses. 

\item Technology maturity: Unlike general-purpose processors, special hardware has a much smaller production volume; the technology available for their production is hence usually several generations behind general-purpose processors. Most AI accelerators, for instance, are based on 28 to 65nm CMOS technology, with a transistor density over 10$\times$ lower than state-of-art mobile CPU or GPU. 

\item Speed: As a consequence of the old technology, special processors run much slower than general-purpose processors do.

\item Eco-system: General-purpose processors have a well-developed eco-system (debugging tools, optimization tools, security measures), which makes the development of high-quality applications much easier than on special processors.

\item Adoption: For all the above reasons, the adoption of a special processor is usually limited to the company that creates it and its few close customers. As a result, an AI application developed for the processor can be adopted by a limited number of devices. 
\end{itemize}

Therefore, whenever mainstream processors can meet the speed and efficiency requirements of an AI application, they should be the preferred device to consider. The common perception that drives the current emphasis in the industry on pursuing special hardware for AI is that mainstream processors are insufficient to meet the real-time requirements. In the rest of this article, we explain why the perception is wrong when compression-compilation co-Design is used, how the principle can be materialized effectively into a practical framework CoCoPIE, and how the future will look like for real-time AI.

\section{Compression-Compilation Co-Design: the Concept}

Compression and compilation are the two key steps in fitting a deep learning model on a hardware for efficient executions. Model compression is a common technique for reducing the size and improving the speed of deep learning models. Compression techniques fall into two categories, {\em pruning} and {\em quantization}. Pruning removes layers or convolution filters or channels, while quantization reduces the precision of parameters (e.g., floating-point to short integer). Compilation refers to the process of generating executable code from a given deep learning model. It, in essence, is a process of mapping the high-level operations in deep learning to the low-level instructions that the underlying hardware supports. The process plays a critical role in optimizing the code for efficient executions. 

The principle of compression-compilation co-design is to design the two components for AI in a hand-in-hand manner. The synergy may exhibit at three levels.

(1) Demands/Preferences Level: At this level, the synergy is on taking the preferences or demands of one component into consideration when designing the other component. An example is that main-stream processors typically prefer code with certain computation patterns; if model compression step can consider that preference, it could create a scenario more amendable for the compilation step to work effectively, as Section~\ref{sec:cocogen} shows.

(2) Perspective/Insight Level: At this level, the synergy is on taking the perspective or insights in the domain of one component when treating the problems in the domain of the other component. An example is the principle of composability or modularity that has been playing an essential role in keeping programming systems and compilations efficient and scalable. Section~\ref{sec:cocotune} will show that when this perspective is introduced into model pruning, large efficiency benefits entail.

(3) Methodology Level: At this level, the synergy is on closely integrating the methodology of the two components together. Section~\ref{sec:cocotune} illustrates this synergy through a compiler framework that automatically generates code to enable a new way of deep learning pruning, which speeds the process by up to 180X. 

All the examples we have mentioned are part of a software framework for Mobile AI named CoCoPIE. We will next give an overview of CoCoPIE, and then uses each of its main components to explain the compression-compilation co-design principle and the significant benefits. 

\section{CoCoPIE}

CoCoPIE stands for \underline{Co}mpression-\underline{Co}mpilation co-design for \underline{P}erformance, \underline{I}ntelligence, and \underline{E}fficiency. It is a software framework that we have recently put together for enabling real-time AI on mainstream end devices.

CoCoPIE holds {\em numerous records on mobile AI}: the first framework that supports all main kinds of DNNs, from CNNs to RNNs, transformer, language models, and so on; the fastest DNN pruning and acceleration framework, up to 180X faster compared with current DNN pruning on other frameworks such as TensorFlow-Lite; making many representative AI applications able to run in real-time on off-the-shelf mobile devices that have been previously regarded possible only with special hardware support; making off-the-shelf mobile devices outperform a number of representative ASIC and FPGA solutions in terms of energy efficiency and/or performance.

CoCoPIE consists of two main components, which both reflect the Compression-Compilation co-design principle. The first component, {\em CoCo-Gen}, generates efficient DNN execution codes via a synergy of pattern-based DNN pruning and pattern-aware code generation. The second component, {\em CoCo-Tune}, dramatically shortens the process in identifying the appropriate set of DNN parameters to prune by a composability-based compiler framework. We next explain each of the two components and how compression-compilation co-design makes them possible.

\subsection{CoCo-Gen: Pattern-based Pruning and Code Generation\protect\footnote{This section is largely based on two published papers~\cite{ma2019pconv,niu2020patdnn}}}
\label{sec:cocogen}

Along with the great success of Deep Neural Networks (DNNs) are the increasingly large model size and complex model structure that require tremendous computation and memory resources to fulfill the {\em real-time} requirement of many key applications. As a mainstream model compression technique, weight pruning is proposed to mitigate this challenge.  Existing pruning however is either incompatible with modern parallel architectures, resulting in long inference latency (e.g., non-structured fine-grained pruning), or subject to significant accuracy degradation (e.g., structured coarse-grained pruning).

\projectname advances the state-of-the-art weight pruning techniques by introducing a new dimension, fine-grained pruning patterns inside the coarse-grained structures, revealing a previously unknown point in the design space. With the higher accuracy enabled by fine-grained pruning patterns, the unique insight is to use the {\em compiler-based code generation} to re-gain and guarantee high hardware efficiency. In other words, our method achieves the best of both worlds, and provides a more favorable option at the levels of theory/algorithm, compiler, and hardware than prior pruning methods. 

\subsubsection{DNN Compression: Challenges and Opportunities}

DNN model compression has been proposed for simultaneously reducing the storage/computation and accelerating inference with minor classification accuracy (or prediction quality) loss. Two important categories of DNN model compression techniques are weight pruning \cite{han2015learning,guo2016dynamic,dai2017nest,mao2017exploring,wen2016learning,he2017channel} and weight quantization \cite{leng2017extremely,park2017weighted,zhou2017incremental,lin2016fixed,wu2016quantized,rastegari2016xnor,hubara2016binarized,courbariaux2015binaryconnect}.

\begin{figure}
    \centering
    \includegraphics[width=0.65 \textwidth]{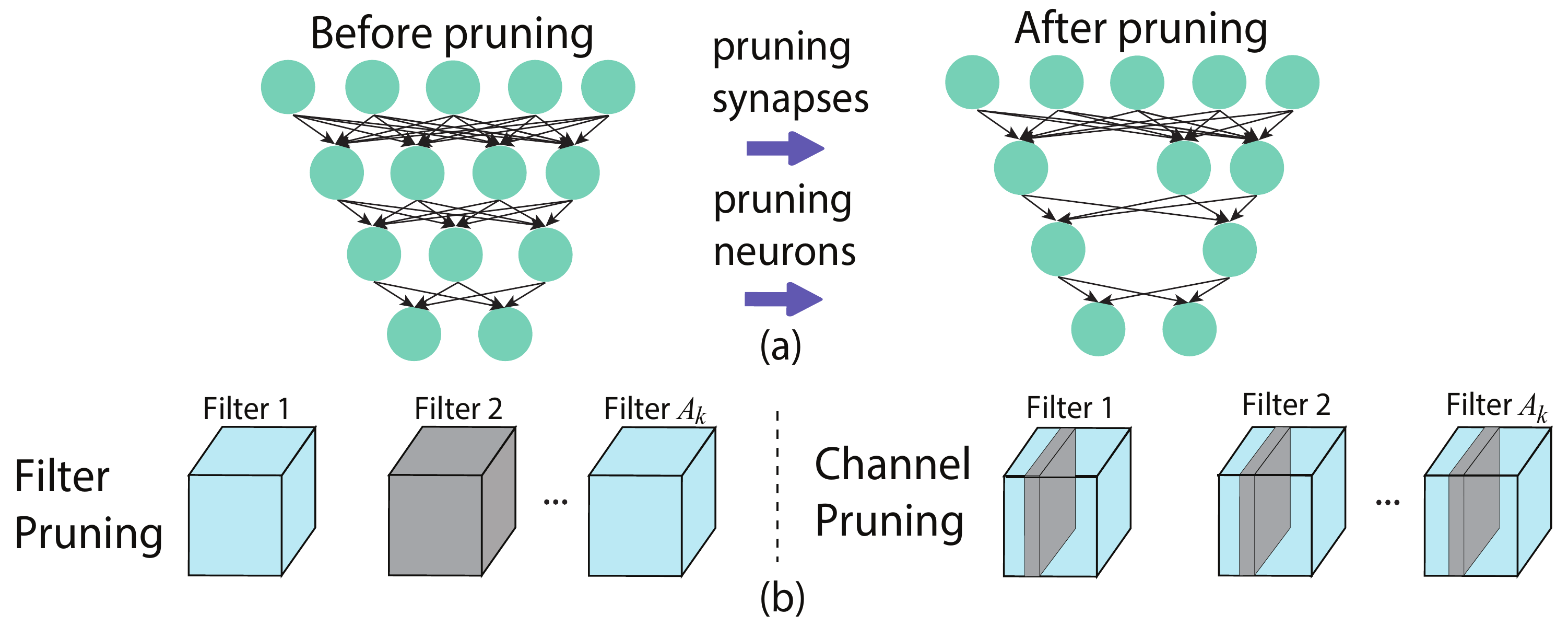}
    \caption{(a) Non-structured weight pruning and {(b) two types of structured weight pruning.}}
    \label{fig:structuredpruning}
\end{figure}

{\em Weight pruning} reduces the redundancy in the number of weights.
As shown in Figure~\ref{fig:structuredpruning}, 
two main approaches of weight pruning are
(1) the general and non-structured pruning; and 
(2) structured pruning. The two approaches produce
irregular and regular compressed DNN models, respectively. 

{\bf Non-Structured Pruning:}
In this method, arbitrary weights can be pruned. It can result in a high pruning rate (i.e., reduction in the number of weights). However, for compiler and code optimization, non-structured pruning incurs several challenges due to the irregularity in computation and memory access. Similarly, for hardware acceleration, since the pruned models are stored in some sparse matrix format with indices, they often lead to performance degradation in GPU and CPU implementations \cite{mao2017exploring,wen2016learning,he2017channel}. 



{\bf Structured Pruning: }
This method can produce regular smaller weight matrices.
{Figure \ref{fig:structuredpruning} (b)} illustrates the representative structured pruning schemes: \emph{filter pruning} and \emph{channel pruning} \cite{wen2016learning}. 
Filter and channel pruning can be considered as equivalent in that pruning a filter in the $k$-th layer is equivalent to pruning the corresponding channel in the $(k+1)$-th layer. Filter/channel pruning is compatible with Winograd algorithm \cite{winograd1980arithmetic,lavin2016fast} 
that has been used to accelerate 
computation of the original DNNs. 
Due to the regular structure, 
the GPU/CPU implementations typically lead to more significant acceleration \cite{mao2017exploring,wen2016learning,he2017channel}.
However, the structured pruning suffers from notable accuracy loss~\cite{wen2016learning,he2017channel}.

{\bf Opportunity:} From the above discussion of non-structured and structured pruning schemes, these two pruning schemes represent two extremes in the design space. In non-structured pruning, any weight can be pruned, and we consider it as a fine-grained method; in structured pruning, the weights of whole filter or channel are pruned, and we consider it as a coarse-grained method. We seek an approach that can offer, or even go beyond, the best of both methods, the high accuracy of non-structured pruning and hardware friendliness of structured ones.

To achieve this goal, we introduce a new dimension, \emph{fine-grained pruning patterns inside the coarse-grained structures}, revealing a previously \emph{unknown} point in the design space.

\subsubsection{Design Philosophy of Pattern-based Pruning}

The proposed \emph{pattern-based pruning} possesses both \emph{flexibility} and \emph{regularity}, and we take a unique approach and leverage compiler optimizations as a bridge between algorithm-level compression and embedded hardware acceleration. The \ul{flexibility} is clearly desirable at the theory and algorithm level, but is also compatible with compiler \emph{code generation} to maximize or maintain both instruction-level and thread-level parallelism. The \ul{regularity} enables another important compiler optimization, \emph{redundant load elimination}, to further improve hardware performance.

The proposed pattern-based pruning techniques consist of \emph{kernel pattern pruning} and \emph{connectivity pruning}. They can be naturally combined in CoCo-Gen, at both algorithm and compiler levels.





\begin{figure}
    \centering
    \includegraphics[width=0.65 \textwidth]{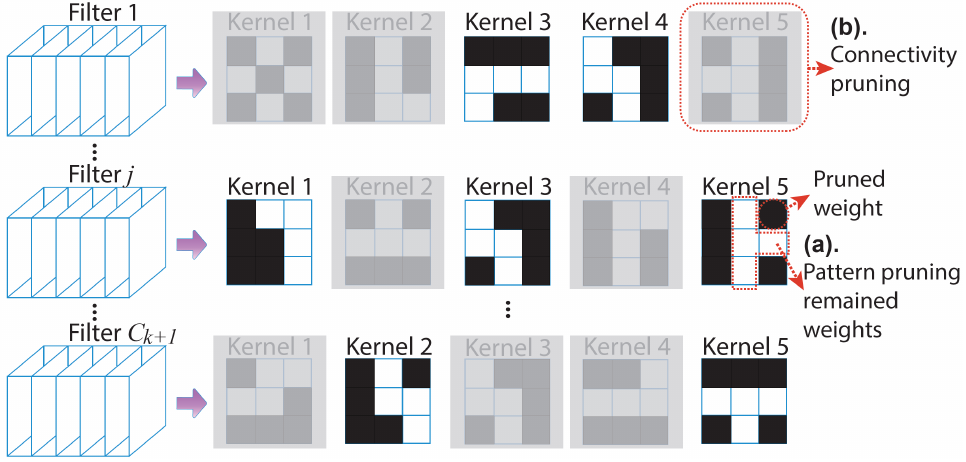}
    \caption{Illustration of (a) kernel pattern pruning on CONV kernels, and (b) connectivity pruning by removing kernels.}
    \label{fig:pattern_connectivity}
\end{figure}

\textbf{Kernel Pattern Pruning} is illustrated in {Figure \ref{fig:pattern_connectivity}}. For each kernel (in a CONV filter), a fixed number of weights are pruned, 
and the remaining weights (white cells) form specific ``patterns''. We define the example in {Figure \ref{fig:pattern_connectivity}} as 4-entry pattern pruning, since every kernel reserves 4 non-zero weights out of the original $3\times 3$ kernel (the most commonly used kernel). We can generalize to other kernel sizes and FC layer. For each kernel, it possesses \emph{flexibility} in choosing among a number of pre-defined patterns.

\ul{At theory and algorithm levels}, it is shown in~\cite{ma2019pconv,li2017pruning,lebedev2016fast} that the desirable kernel shape has certain patterns to match the connection structure in human visual systems, instead of a square shape. The selection of appropriate pattern for each kernel can be achieved by extending ADMM-based framework. As shown in~\cite{ma2019pconv}, we achieve accuracy enhancement in all representative DNNs in our testing. \ul{At compiler level}, the known patterns allow a compiler to \emph{re-order and generate codes} at filter level and kernel level to group kernels with the same pattern for consecutive executions, thereby maximizing instruction-level parallelism. \ul{At hardware level}, 4-entry patterns perfectly fit the SIMD architecture in embedded processors, for both CPUs and GPUs.


\begin{figure}[t]
    \centering
    \includegraphics[width=0.65 \textwidth]{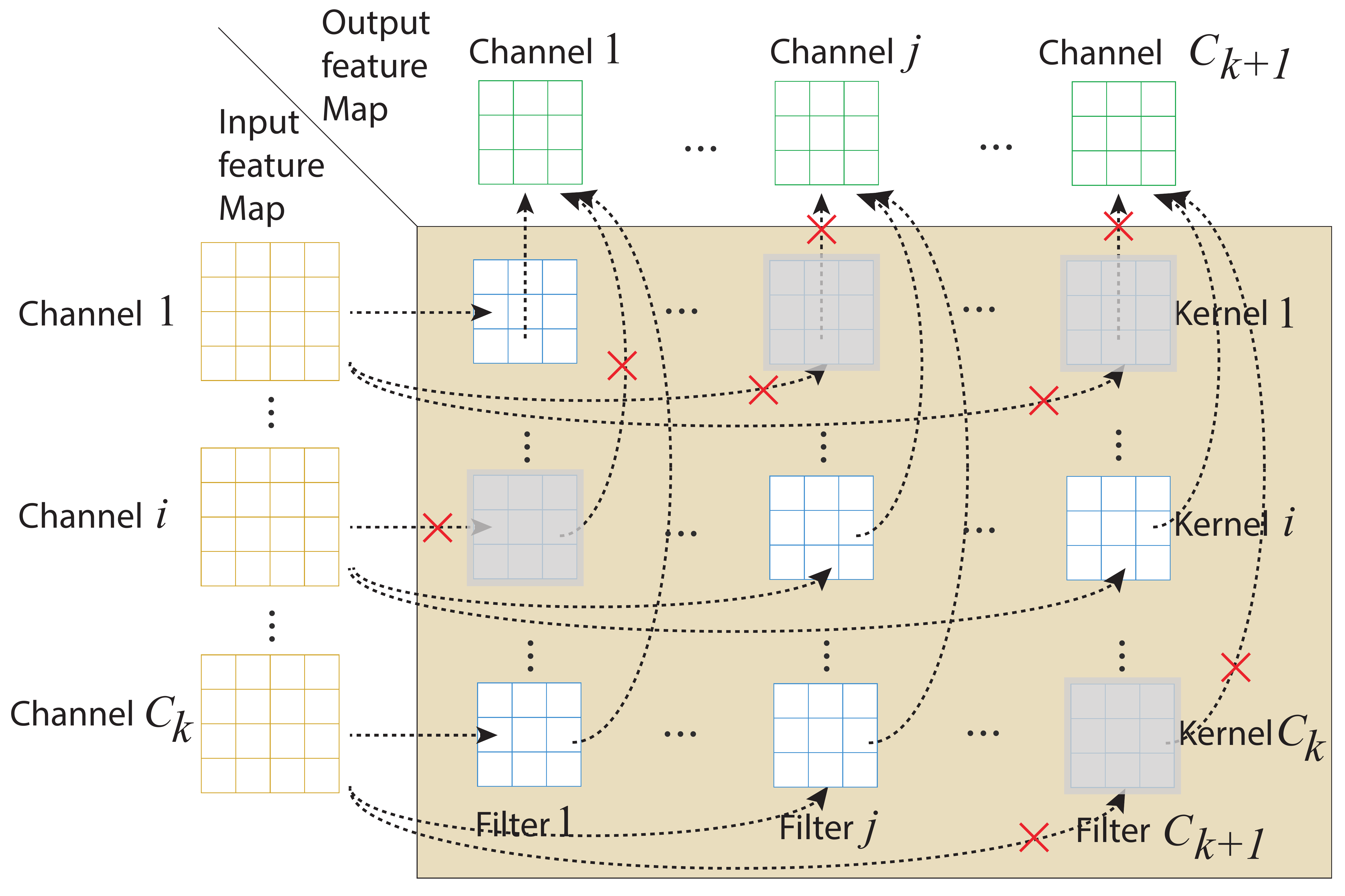}
    \caption{Illustration of connectivity pruning.}
    \label{fig:connectivity}
\end{figure}

\textbf{Connectivity Pruning} is illustrated in {Figure \ref{fig:connectivity}}.
The key insight is to \emph{cut the connections} between certain input and output channels, which is equivalent to the removal of corresponding kernels. The method is proposed to achieve higher weight pruning/acceleration rates in combination with kernel pattern pruning.

\ul{At theory and algorithm levels}, connectivity pruning matches the desirability of locality in layerwise computations inspired by human visual systems~\cite{yamins2016using,yamins2014performance}. It is more flexible than filter/channel pruning and achieves higher accuracy. \ul{At compiler and hardware levels}, removed kernels and associated computations are grouped by compiler using the \emph{re-ordering} capability without affecting the other computations, thereby maintaining parallelism degree.


\begin{table}[t]
\vspace{-5mm}
\caption{Qualitative comparison of different pruning schemes on accuracy and speedup under the same pruning rate.}\label{tab:prunecompare}
\centering
\includegraphics[width =  0.65\linewidth]{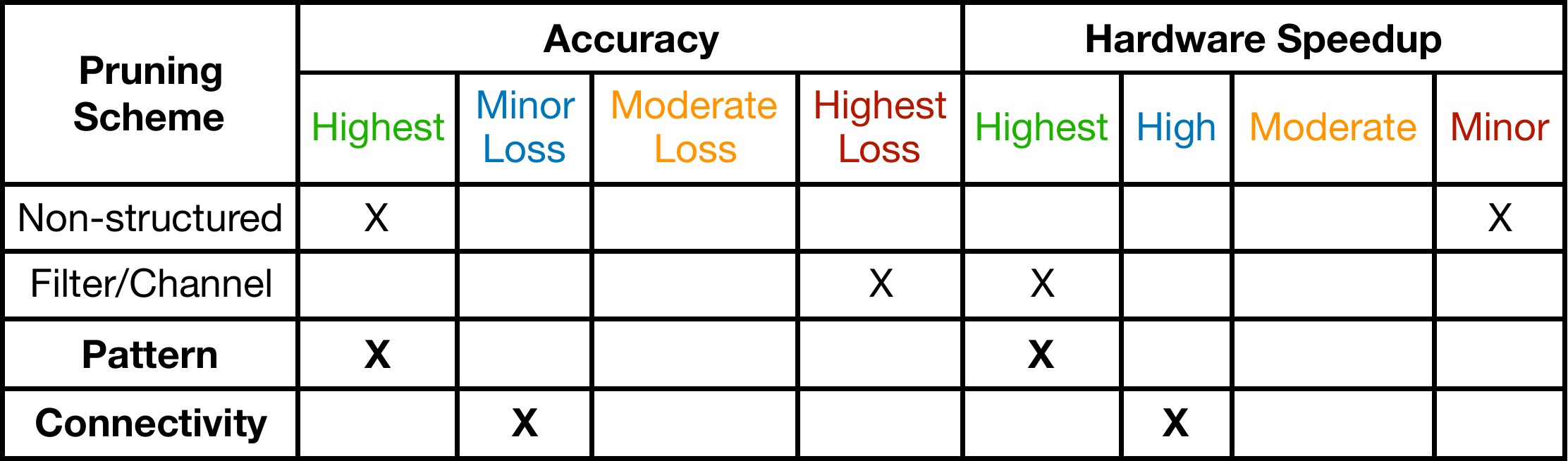}
\end{table}


\subsubsection{Internal Mechanisms of \projectname}

\begin{figure}[t]
    \centering
    \includegraphics[width=1 \textwidth]{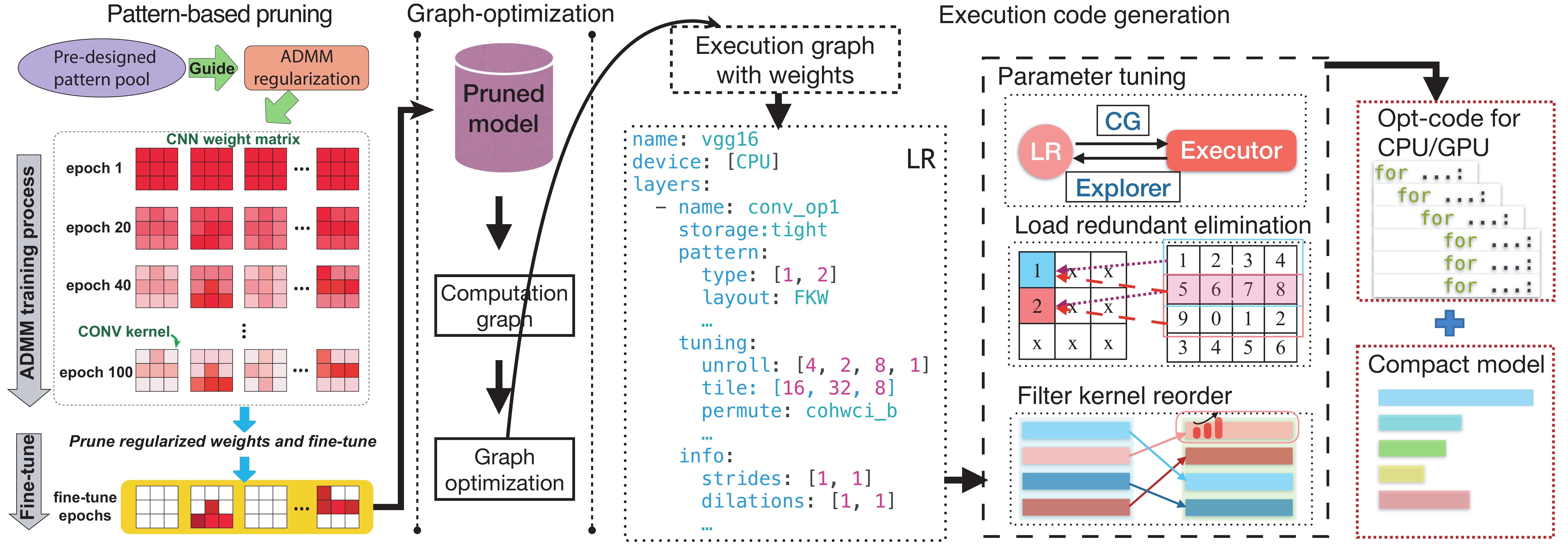}
    \caption{Overview of \projectname acceleration framework.}
    \label{fig:system-overview}
\end{figure}



Figure~\ref{fig:system-overview} 
shows the overview of \projectname which consists of
two stages:
(1) {\em pattern-based training stage}, which
performs kernel pattern and connectivity pruning
with an extended ADMM solution framework. 
(2) {\em execution code generation stage}, which
performs multiple effective optimizations based
on the patterns. 
Similar to TVM \cite{chen2018tvm}, \projectname converts DNN models into computational graphs and applies multiple graph-based optimizations. 
Based on these optimizations, we focus on
layerwise design and optimization including 
a high-level and fine-grained DNN layerwise representation (LR), filter kernel reorder, load redundancy eliminations, and automatic parameter tuning. All of these designs and optimizations are general, 
and applicable to both mobile CPUs and GPUs.
The second stage generates optimized execution codes as well as DNN models with weights stored in a novel compact format. 
We briefly explain each component as follows. 


{\bf Pattern-based training stage} performs effective kernel pattern and connectivity pruning in the training phase, in order to achieve the highest pruning (acceleration) rate without accuracy loss. First, we design a set of patterns to select for each kernel. Then we perform pattern pruning based on the designed pattern set and connectivity pruning, using an extended ADMM-based method.

{\bf Fine-grained DNN layerwise representation (LR)} provides a high-level representation to enable our general optimization on DNN models from various resources. 
This LR captures more extensive information of each DNN layer compared with the TVM's IR. 
In particular, it includes the pattern and tuning related information. The compiler optimizations rely on a series of improvements on this LR to generate the compact model and optimized execution codes.

{\bf Filter kernel reorder} addresses two 
challenges of pattern-based pruning---heavy control-flow instructions, and thread divergence and load imbalance---by grouping the filters with similar lengths and patterns together. 
Because of the relatively limited number of patterns, the kernels with similar patterns can be organized together through proper filter kernel reordering, thereby significantly reducing the control-flow instructions and improving the {\em instruction-level parallelism}. Moreover, if different threads process different filters, thread divergence and load imbalance issues are properly resolved because the kernels in each filter have similar computation workload, thereby enhancing {\em thread-level parallelism}.


{\bf Compressed weight storage} is specifically designed for our kernel pattern and connectivity pruning. Together with filter kernel reorder, this compact data structure yields much better compression rates than the conventional CSR (compressed sparse row) format.

{\bf Load redundancy elimination} addresses the
poor memory performance challenge 
of pattern-based pruning 
by exploring two novel {\em register-level} load redundancy opportunities during the kernel execution code generation.
It is crucial, especially when the data movements between memory and cache have already been optimized with advanced tiling techniques. 

{\bf Parameter auto-tuning} specifically tests on different configurations of the key performance parameters, including strategies of placing data on various GPU memories, different tiling sizes, and loop permutations for each DNN layer on each processing unit.

In sum, allowing compilers to treat 
pruned kernels as special patterns, 
our approach not only achieves high pruning rate
with high accuracy, 
but also effectively converts the patterns into performance 
improvements for their hardware friendly properties. 
%
As shown in {Table \ref{tab:prunecompare}}, \projectname can achieve the benefits of both non-structured and structured pruning. 
It illustrates the co-design principle:
the multi-level cache memory hierarchy provides
sufficient hardware supports to hide memory
access latency and explore locality, and the SIMD units offers vector/parallel computing capability, the potential of which is unleashed more effectively when the compilation and compression processes are co-designed synergistically. 

\subsubsection{Evaluation and Demos}

\begin{figure}[t]
    \centering
        \subfloat[ImageNet-CPU]{
            \includegraphics[width=0.235\textwidth]{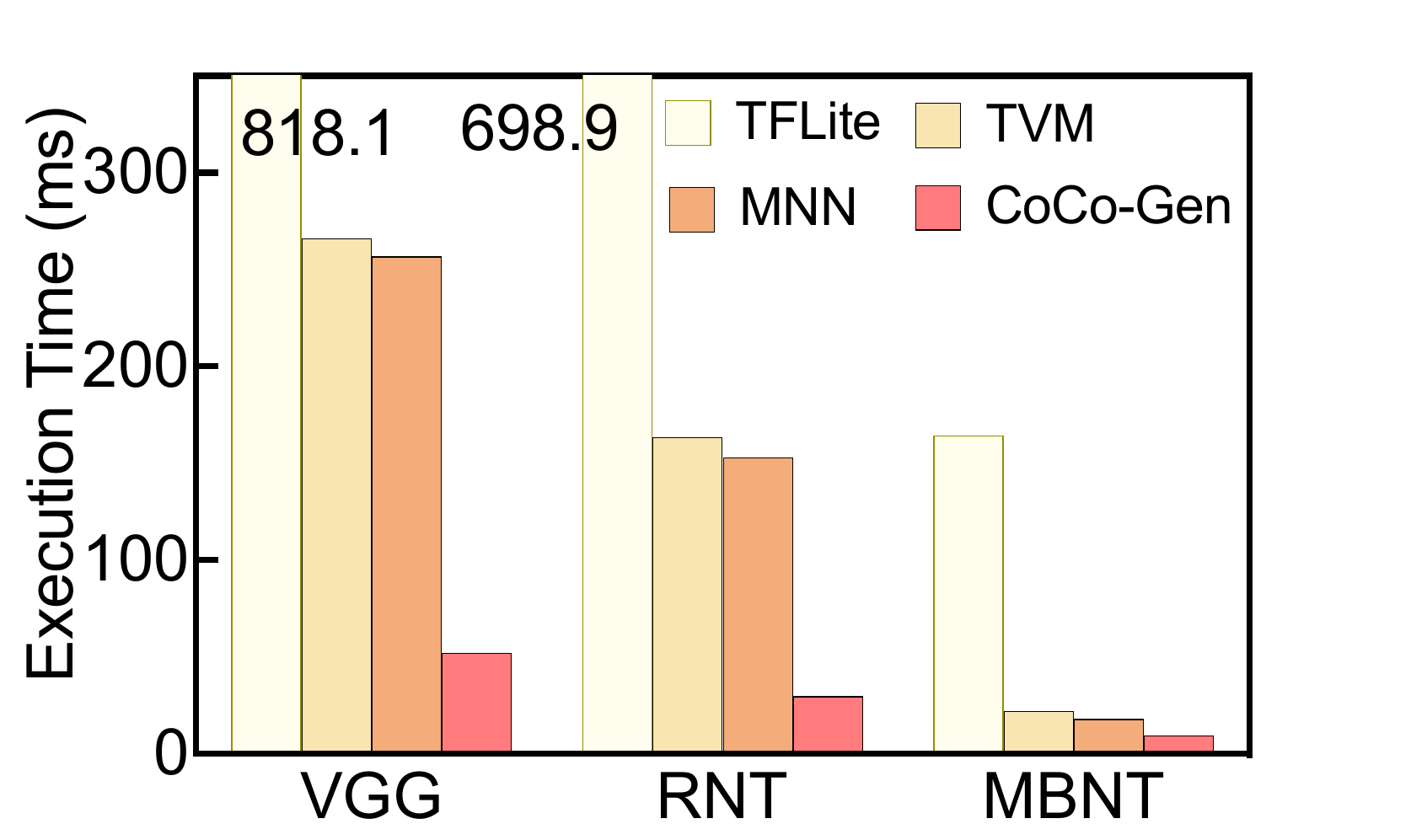}
        }
        \subfloat[CIFAR-10-CPU]{
            \includegraphics[width=0.25\textwidth]{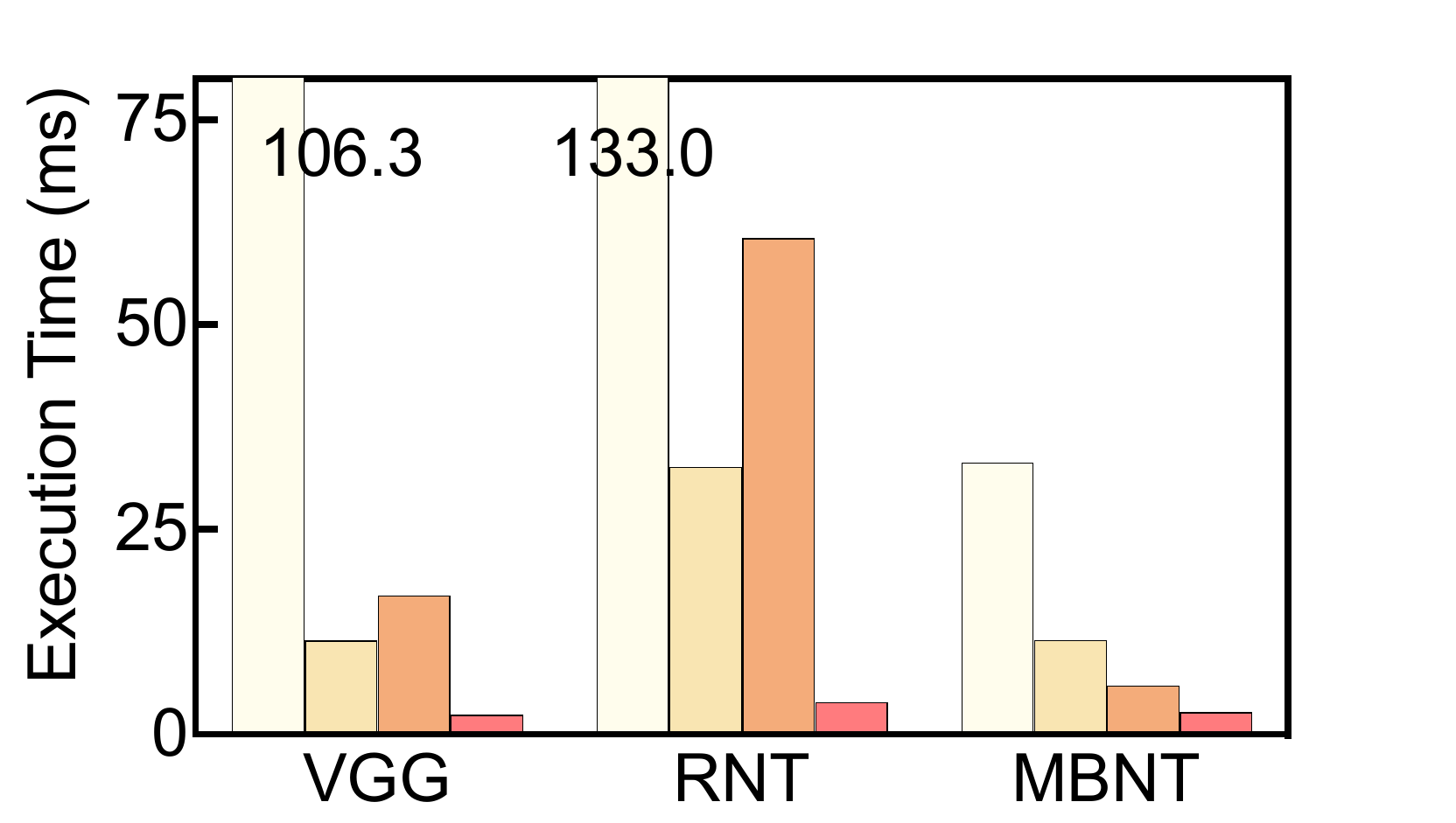}
        }
        \subfloat[ImageNet-GPU]{
            \includegraphics[width=0.25\textwidth]{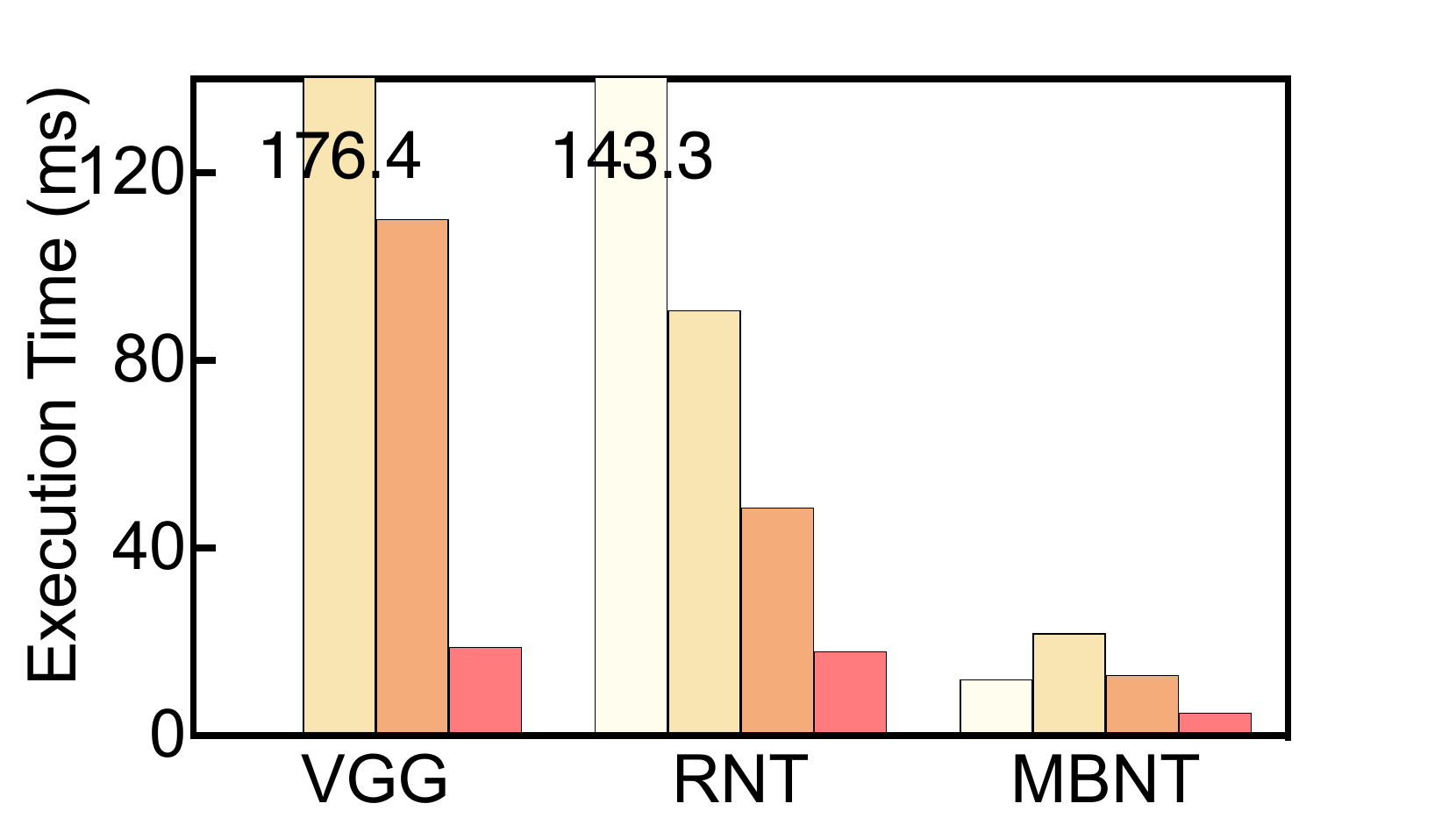}
        }
        \subfloat[CIFAR-10-GPU]{
            \includegraphics[width=0.25\textwidth]{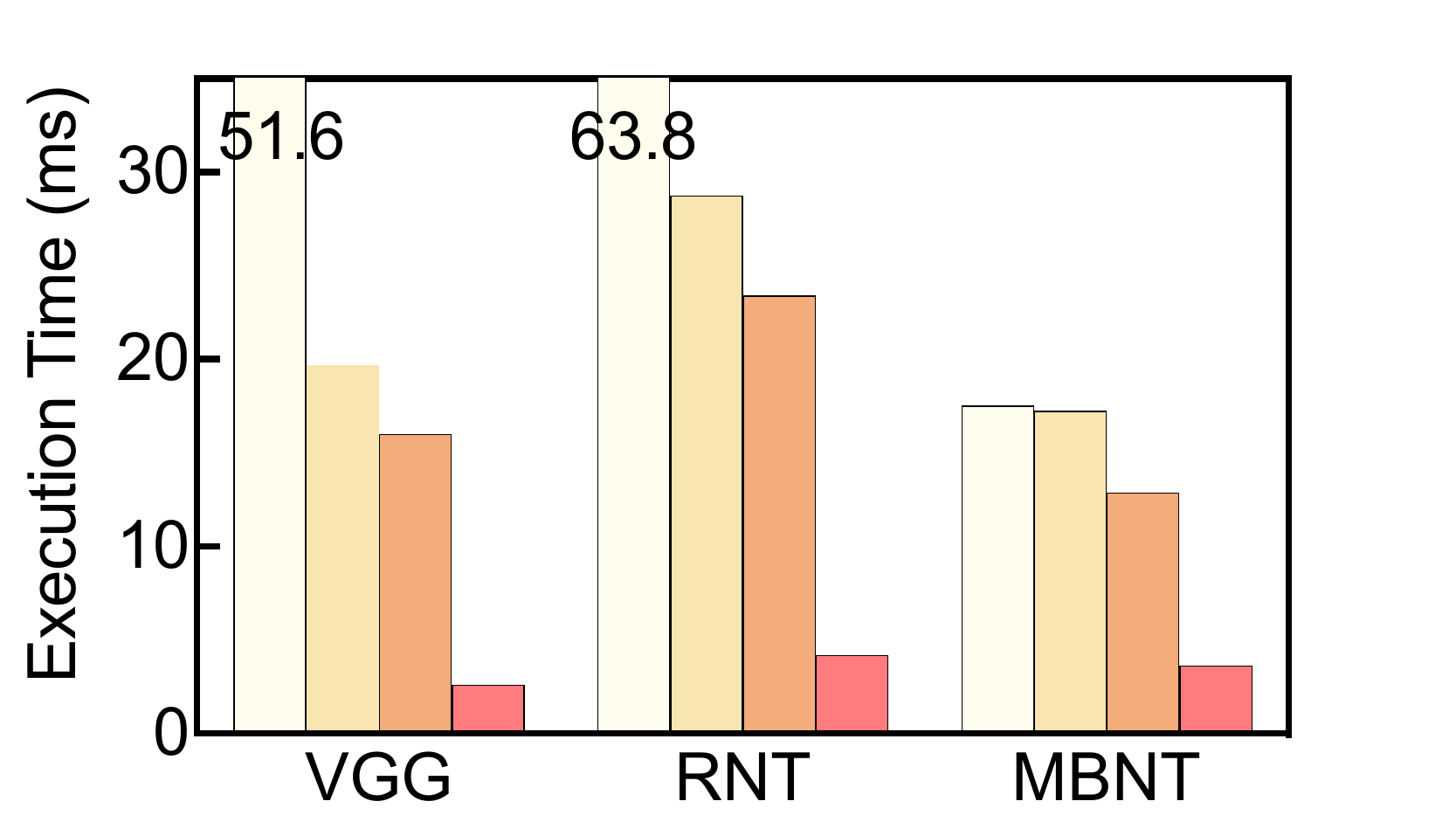}
        }
        \caption{Performance comparison: x-axis: different trained DNN models; y-axis: average DNN inference execution time on a single input.}
    \label{fig:eva_overview_performance}
\end{figure}

 {\bf Results on DNNs:} We evaluate \projectname on a Samsung Galaxy S10 cell phone with the latest Qualcomm Snapdragon 855 mobile platform that consists of a Qualcomm Kryo 485 Octa-core CPU and a Qualcomm Adreno 640 GPU. Figure~\ref{fig:eva_overview_performance} shows the CPU and GPU performance of \projectname compared to TFLite~\cite{TensorFlow-Lite}, TVM~\cite{chen2018tvm}, and MNN~\cite{Ali-MNN} on six representative DNNs, VGG-16 (VGG), ResNet-50 (RNT), and MobileNet-V2 (MBNT) trained on two datasets, ImageNet and CIFAR-10.  \projectname outperforms all other frameworks for all cases. 
On CPU, \projectname achieves 
$12\times$ to $44.5\times$ speedup over TFLite, $2.3\times$ to $8.1\times$ over TVM, and $1.9\times$ to $15.5\times$ over MNN, 
respectively. 
On GPU, \projectname achieves 
$2.5\times$ to $20\times$, $4.1\times$ to $11.4\times$, and $2.5\times$ to $6.2\times$ 
speedup over TFLite, TVM, and MNN, respectively\footnote{TFLite does not support executing VGG on ImageNet data set on GPU due to its too large memory footprint.}. For the largest DNN (VGG) and largest data set (ImageNet), \projectname completes CONV layers on a single input  within 
18.9 ms on GPU, meeting the real-time requirement (usually 30 frames/sec, 
i.e., 33 ms/frame).


\begin{figure}[tb]    
 \centering
 \scalebox{1.3}[1.3]{
\begin{tabular}{p{0.17in}p{1.3in}p{1.3in}p{1.3in}}
& \parbox{1.3in}{\centering \footnotesize style transfer} &  
\parbox{1.3in}{\centering \footnotesize coloring}
&  
\parbox{1.3in}{\centering \footnotesize super resolution}
\\
 \rotatebox{90}{\parbox{1.3in}{\centering \footnotesize original image }}
 &   
\includegraphics[width=1.3in,height=1.3in]{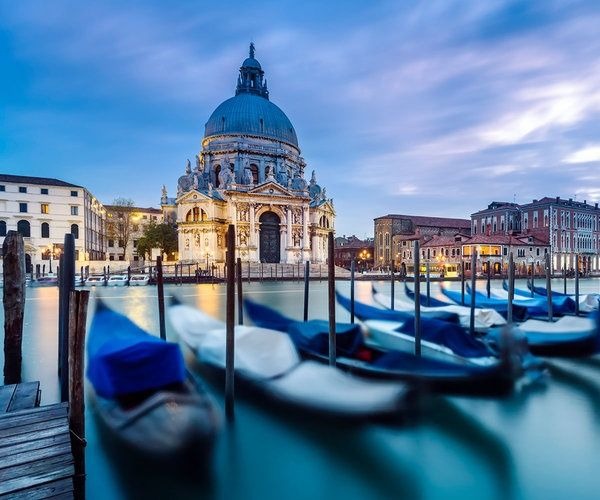}& 
\includegraphics[width=1.3in,height=1.3in]{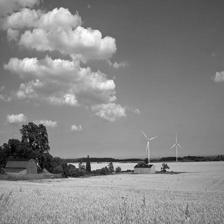}& 
\includegraphics[width=1.3in,height=1.3in]{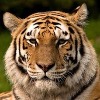}
\\
 \rotatebox{90}{\parbox{1.3in}{\centering \footnotesize application output }}
&   
\includegraphics[width=1.3in,height=1.3in]{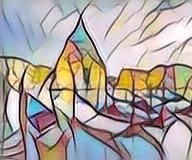} &  
\includegraphics[width=1.3in,height=1.3in]{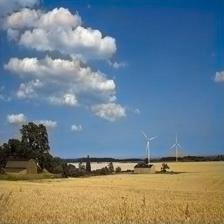} &  
\includegraphics[width=1.3in,height=1.3in]{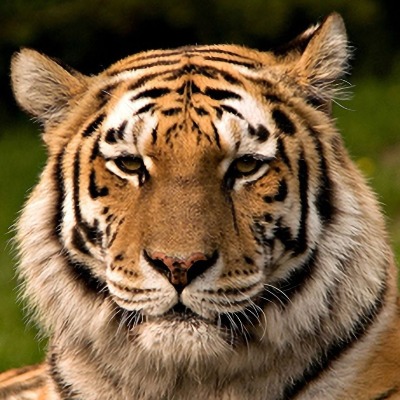}
\end{tabular}}
\caption{\footnotesize{ Examples of  style transfer, coloring, and super resolution implemented on our mobile device.
}} \label{fig-examples}
\end{figure}


{\bf Real Application Demos:} We also demonstrate the efficacy of \projectname through three interesting and key DNN applications, style transfer~\cite{gatys2016image}, DNN coloring~\cite{2925974}, and super resolution~\cite{dong2014learning}. The style transfer model is based on a generative network~\cite{zhang2017multistyle} trained on Microsoft COCO~\cite{lin2014microsoft}.  DNN coloring uses the Places scene~\cite{zhou2014learning} dataset to train a novel architecture that can jointly extract  and  fuse global and local features  to perform the final colorization. The super resolution model mainly utilizes residual blocks with wider activation and linear low-rank convolution~\cite{yu2018wide} trained on the DIV2K~\cite{timofte2017ntire} dataset. With structured pruning and compiler optimization, we implement the models on a Samsung Galaxy S10 mobile phone. We demonstrate that our implementations are able to achieve real-time inference on off-the-shelf mobile device with video demos.

Figure~\ref{fig-examples} shows sample input and output of three applications.
\projectname optimization accelerates the inference with speedups of $4.2\times$, $3.6\times$, and $3.7\times$ for style transfer, coloring and super resolution, respectively. 
 These results demonstrate that our optimized implementation generates satisfying output with high speed on mobile devices.
 More specifically, all inference can complete within 75 ms, showing the possibility of achieving real-time executions of complex DNN applications on such main-stream devices without special hardware. Please find more video demos at our YouTube channel\footnote{\url{ www.youtube.com/channel/UCCKVDtg2eheRTEuqIJ5cD8A/}.
}. 

\subsubsection{Outperforming Existing ASIC and FPGA Solutions in Performance/Energy Efficiency}

\begin{figure}[tb]    
 \centering
 \scalebox{1.0}[1.0]{
\begin{tabular}{p{1.83in}p{1.83in}p{1.83in}}
\\
\includegraphics[width=1.83in,height=1.6in]{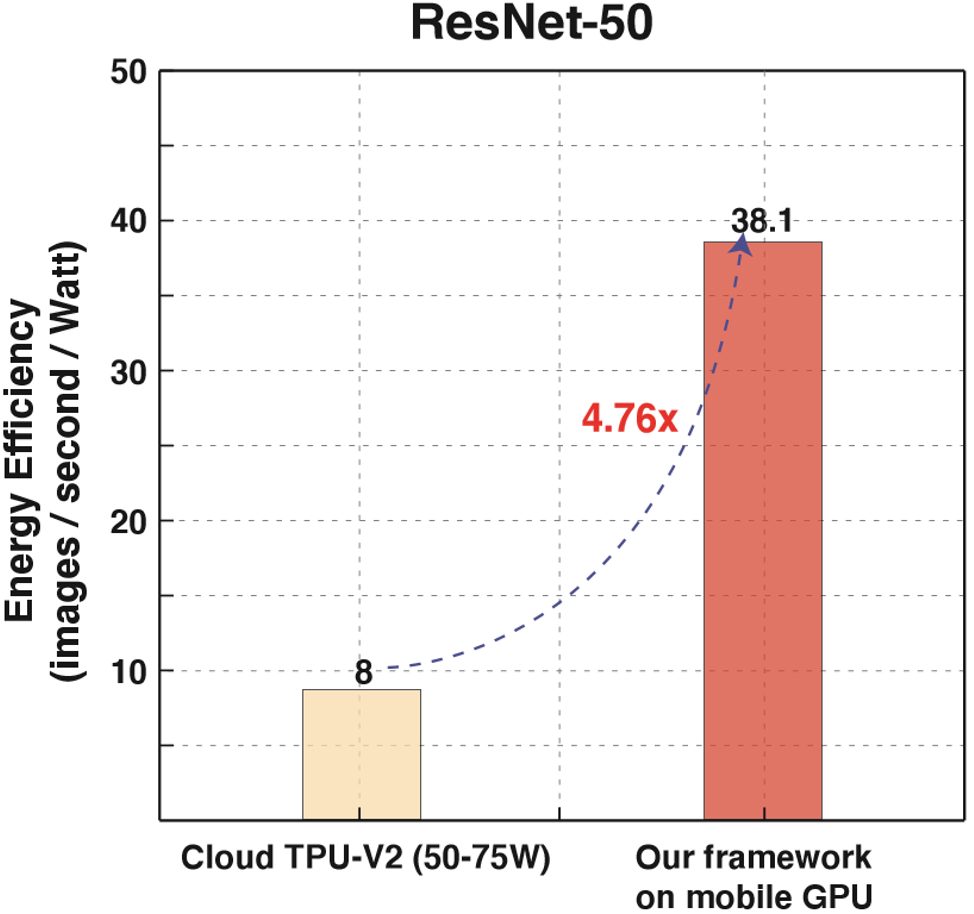}& 
\includegraphics[width=1.83in,height=1.6in]{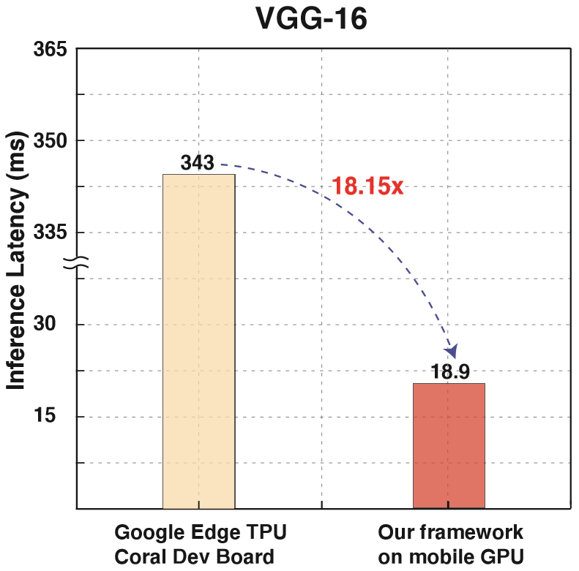}& 
\includegraphics[width=1.83in,height=1.55in]{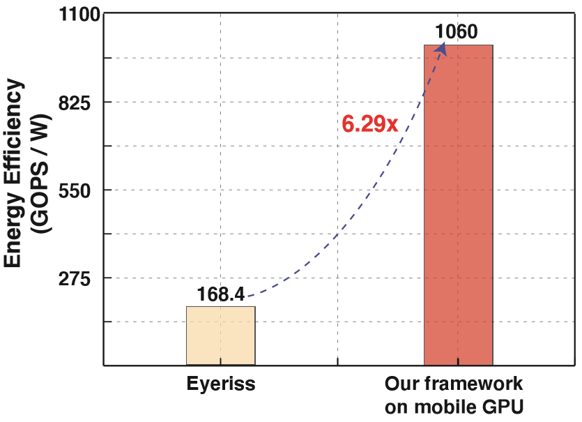} 
\\
\multicolumn{2}{c}{(a) Google TPU}   &  
\multicolumn{1}{c}{(b)   Eyeriss} 
\\
\\
\includegraphics[width=1.83in,height=1.6in]{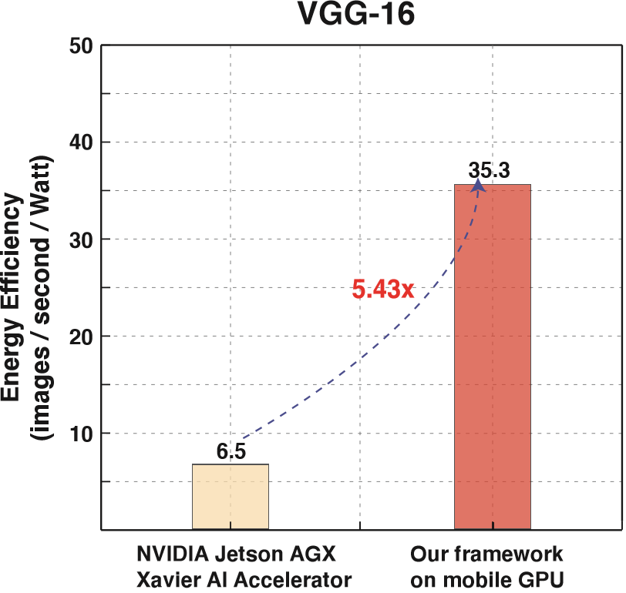}&
\includegraphics[width=1.83in,height=1.6in]{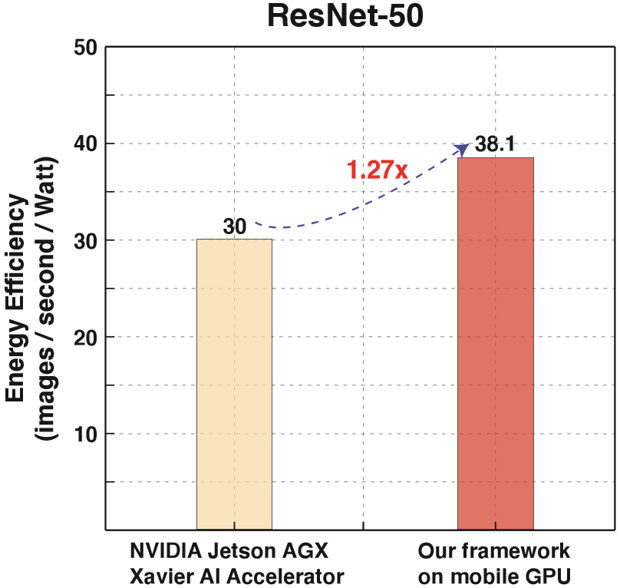}&
\includegraphics[width=1.83in,height=1.6in]{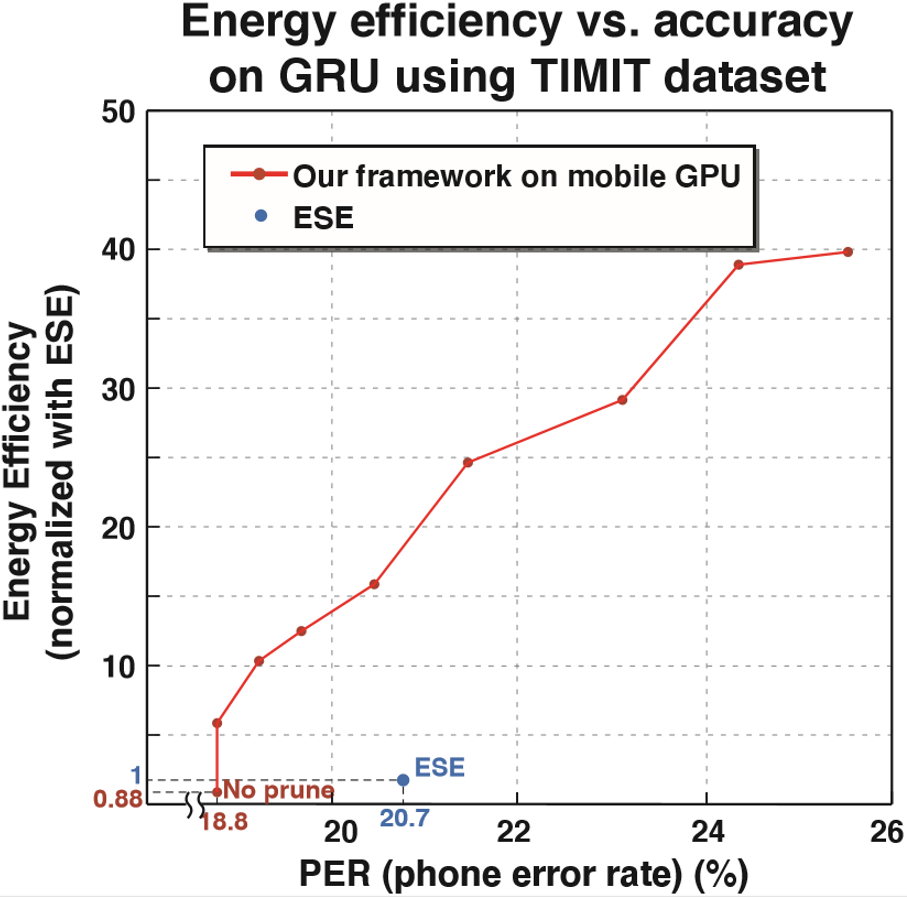}
\\
\multicolumn{2}{c}{(c)  NVIDIA Xavier} &   \multicolumn{1}{c}{ (d)  ESE} 
\end{tabular}}
\caption{\footnotesize{ Comparison with existing ASIC and FPGA solutions. (a) Comparison of energy efficiency and inference latency with Google cloud TPU. (b) Comparison of energy efficiency  with  Eyeriss. (c) Comparison of energy efficiency  with  NVIDIA Jetson AGX Xavier. (d) Comparison of energy efficiency  with  FPGA solution ESE. 
}} \label{fig-comparison-with-ASIC-FPGA}
\end{figure}

Using the CoCo-Gen on off-the-shelf general-purpose mobile device (e.g., the Samsung Galaxy S10 smartphone), we consistently outperform a number of ASIC and FPGA solutions in performance and energy efficiency. Figure \ref{fig-comparison-with-ASIC-FPGA} demonstrates (i) the comparison results on performance and energy efficiency with special ASIC hardware including Google's cloud TPU-V2 and edge TPU~\cite{googletpu}, NVIDIA Jetson AGX Xavier, Eyeriss~\cite{isscc_2016_chen_eyeriss}, etc., and (ii) comparison results on accuracy and energy efficiency with the FPGA solution ESE \cite{han2017ese} (FPGA 2017 Best Paper Award) from DeePhi. The comparisons are on the same network models, and weight quantization is not applied in CoCo-Gen solution (Eyeriss and ESE use 12-bit fixed-point quantizations).

We can clearly observe that our CoCo-Gen solution on general-purpose mobile device consistently outperforms representative ASIC/FPGA solutions in terms of energy efficiency. This unusual phenomenon is attributed to three reasons: (i) smartphone itself has ultra-high energy efficiency. Smartphone computing chips are built using the most advanced technology (e.g., 7nm, 11nm technology) and are the key driving force of technology advancement, while FPGA/ASIC solutions are based on 28nm or 40nm technologies which are inherently less energy-efficient. Also ARM (for mobile CPU) and Qualcomm (for mobile GPU) are especially proficient in high-efficiency circuit/system designs. (ii) while prior mobile compiler framework has limited support on different neural networks (e.g., not supporting RNNs or large-scale DNNs), our CoCo-Gen compiler can support all of the major types of neural networks, thereby unleashing the full potential of mobile devices. (iii) the unique benefit of compression-compilation co-design. Additionally, it can be observed that CoCo-Gen achieves consistently high performance on different DNN benchmarks thanks to the high flexibility of software-based solution. In contrast, current ASIC/FPGA solutions are optimized for certain DNN types/sizes, thereby lacking generality. For example, edge TPU is optimized for small-scale DNNs while cloud TPU is optimized for large-scale ones.

\vspace{.1in}
The studies reported in this part have shown that effective pruning coupled with pattern-based compilation can bring large performance benefits for DNNs. But finding out what is the best set of filters or connectivities to prune can be extremely time consuming. For a DNN with $W$ filters, the entire configuration space of pruned network can be as large as $2^{|W|}$, even if only filter pruning is considered (adding pattern variations would worsen the complexity further). It often takes hours to evaluate just one configuration (i.e., training the pruned network and then testing it). We next show that the process can be shortened dramatically through a compiler-based framework, CoCo-Tune. It offers an example on the benefits of combining compiler perspective and support with DNN compression. 

\subsection{CoCo-Tune: A Compiler Framework for Fast Pruning\protect\footnote{This section is largely based on a published paper~\cite{Guan+:PLDI2019}.}}
\label{sec:cocotune}

CoCo-Tune is a compiler-based framework designed for shortening the time needed for CNN pruning in order to remove the major barrier for timely solution delivery in Artificial Intelligence (AI) product development, especially on mobile devices. Prior efforts on speeding up pruning have been, however, mostly focused on only the compression domain~\cite{li2016pruning,hu2016network,molchanov2016pruning,luo2017thinet,he2018adc}. They leverage DNN algorithm-level knowledge to reduce the enormous configuration space to a smaller space (called {\em promising subspace}) that is likely to contain a good solution, and then evaluate these remaining configurations to find the best.

Although these prior methods help mitigate the problem, network pruning remains a time-consuming process. One reason is that, despite their effectiveness, no prior techniques can guarantee the inclusion of the desirable configuration in a much reduced subspace. As a result, to decrease the risk of missing the desirable configuration, practitioners often end up with a still quite large subspace of network configurations that takes days for many machines to explore. It is also quite often that modifications need to make to the CNN models, datasets, or hardware settings throughout the development process of an AI product; each of the changes could make the result of a CNN pruning obsolete and call for a rerun of the entire pruning process. 

This study distinctively examines the problem from the programming systems perspective. Specifically, rather than improving the attainment of promising subspace as all prior work focuses on, we try to drastically speed up the evaluations of the remaining configurations in the promising subspace through cross-network {\em computation reuse} via a compiler-based framework, a direction that has never been explored before. 

We achieve the goal through three-fold innovations. First, we empirically uncover the existence of {\em composability} in the training of a collection of pruned CNN models, and reveal the opportunity that the composability creates for saving computations in CNN pruning. The basic observation that leads to this finding is that two CNN networks in the promising subspace often differ in only some layers.
In the current CNN pruning methods, the two networks are both trained from scratch and then tested for accuracy. A question asked in this work is whether the training results of the common layers can be reused across networks to save some training time. More generally, we view the networks in a promising subspace as compositions of a set of building blocks (a {\em block} is a sequence of CNN layers). The question is if we first pre-train (some of) these building blocks and then assemble them into the to-be-explored networks, can we shorten the evaluations of these networks and the overall pruning process? Through a set of experiments, we empirically validate the hypothesis, based on which, we propose {\em composability-based CNN pruning} to capture the idea of reusing pre-trained blocks for pruning. 

Second, we propose a novel {\em hierarchical compression-based algorithm}, which, for a given CNN and promising subspace, efficiently identifies the set of blocks to pre-train to maximize the benefits of computation reuse. We prove that identifying the optimal set of blocks to pre-train is NP-hard. Our proposed algorithm provides a linear-time heuristic solution by applying Sequitur~\cite{nevill1997identifying}, a hierarchical compression algorithm, to the CNN configurations in the promising subspace.

Finally, based on all those findings, we developed CoCo-Tune, the first compiler-based framework that, for an arbitrary CNN (in Caffe Prototxt format) and other inputs, automatically generates TensorFlow code to build Teacher-Student learning structures to materialize  composability-based CNN pruning. 

\subsubsection{Composability-Based CNN Pruning: Idea and Challenges}
\label{sec:comp}

The fundamental reason for CoCo-Tune to produce large speedups for CNN pruning is its effective capitalization of computation reuse based on the {\em composability in CNN pruning} that is empirically unveiled in this study. Two pruned networks in a promising subspace often differ in only some of the layers. The basic idea of {\em composability-based CNN pruning} is to reuse the training results of the common layers across the pruned networks. Although the idea may look straightforward, to our best knowledge, no prior CNN pruning work has employed such reuse, probably due to a series of open questions and challenges:

\begin{itemize}
\item First, there are bi-directional data dependencies among the layers of a CNN. In CNN training, for an input image, there is a forward propagation that uses a lower layer's output, which is called {\bf activation maps}, to compute the activation maps of a higher layer; it is followed by a backward propagation, which updates the weights of a lower layer based on the errors computed with the higher layer's activation maps. As a result of the bi-directional dependencies, even just one-layer differences between two networks could cause very different weights to be produced for a common (either higher or lower) layer in the two networks. Therefore, it remains unclear whether the training results of a common layer could help with the training of different networks. 

\item Second, if a pre-trained layer could help, it is an open question how to maximize the benefits. A pre-trained sequence of consecutive layers may have a larger impact than a single pre-trained layer does on the whole network, but it may also take more time to produce and has fewer chances to be reused. How to determine which sets of layers or sequences of layers to pre-train to maximize the gains has not been explored before. 

\item Third, how to pre-train just a piece of a CNN? The standard CNN back propagation training algorithm uses input labels as the ground truth to compute errors of the current network configurations and adjust the weights. If we just want to train a piece of a CNN, what ground truth should we use? What software architecture should be built to do the pre-training and do it efficiently?
 
\item Fourth, existing DNN frameworks support only the standard DNN training and inference. Users have to write code to do CNN pruning themselves, which is already complicated for general programmers. It would add even more challenges to ask them to additionally write the code to pre-train CNN pieces, and then reuse the results during the evaluations of the networks.  
\end{itemize}

For the first question, we conduct a series of experiments on 16 large CNNs (four popular CNN models trained on four datasets). We briefly state the key observations here (see~\cite{Guan+:PLDI2019} for details). Pre-trained layers bring a network to a much improved starting setting, making the initial accuracies of the network 50-90\% higher than the network without pre-trained layers. That leads to 30-100\% savings of the training time of the network. Moreover, it helps the network converge to a significantly higher level of accuracy (by 1\%-4\%). These findings empirically confirm the potential of {\em composability-based CNN pruning}. 

To effectively materialize the potential, we have to address the other three challenges. CoCo-Tune offers the solution.

\subsubsection{CoCo-Tune Framework}
\label{sec:overview}

\begin{wrapfigure}{r}{.5\textwidth}
    \centering
    \includegraphics[width=.5\textwidth]{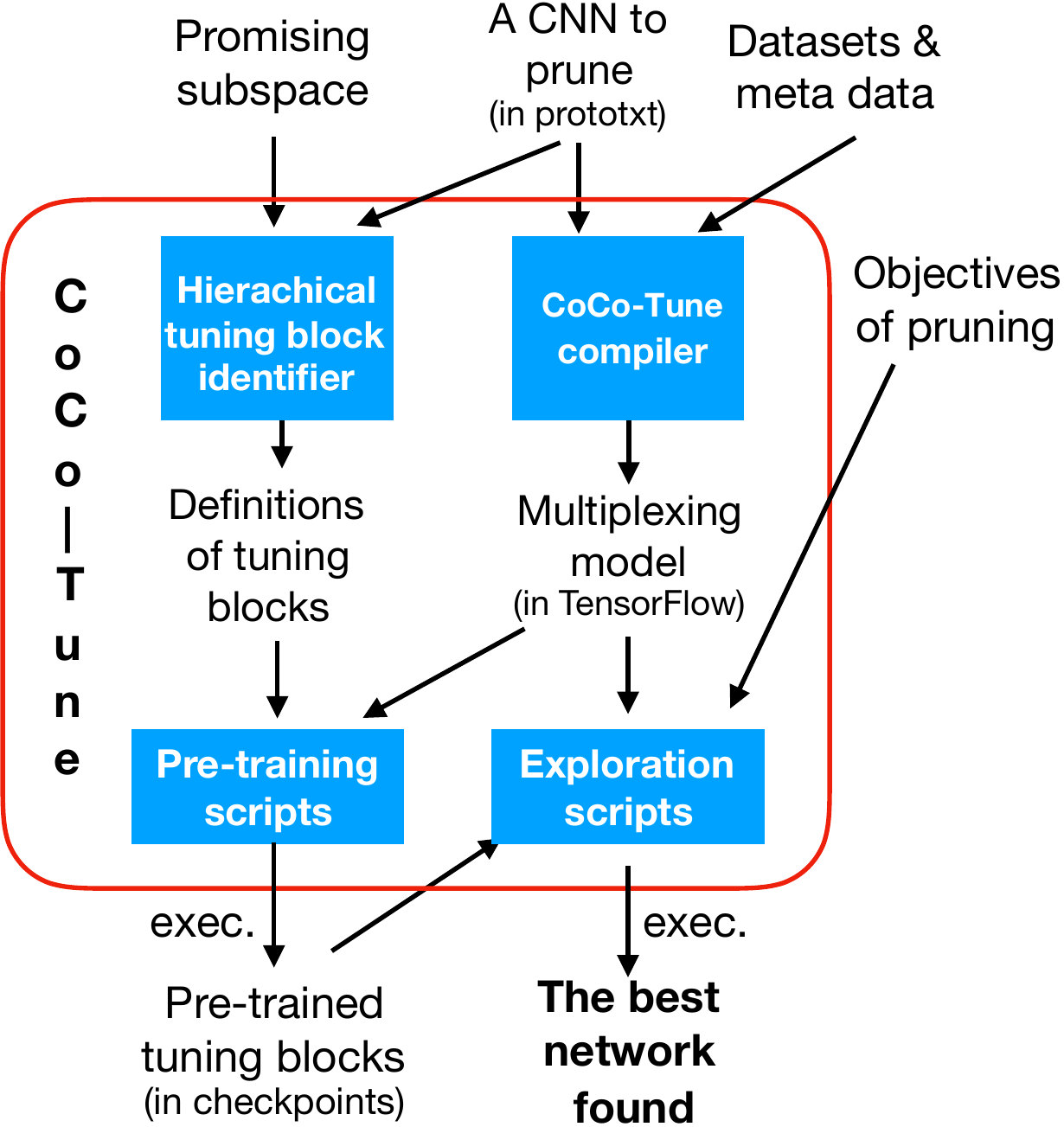}
    \caption{Overview of CoCo-Tune Framework.}
    \label{fig:overview}
\end{wrapfigure}

CoCo-Tune is a software framework that automatically enables composability-based CNN pruning. As Figure~\ref{fig:overview} shows, its input has four parts: 

\begin{itemize}
    \item The to-be-pruned CNN model, written in Caffe Prototxt (with a minor extension), which is a user-friendly text format (from Caffe) for CNN model specifications~\cite{jia2014caffe}. 
    \item The promising subspace that contains the set of pruned networks configurations worth exploring. The subspace may come from the user or some third-party tools that reduce the configuration space for CNN pruning~\cite{hoos2011automated,he2018adc,ashok2017n2n}. 
    
    \item The dataset for training and testing, along with some meta data on the training (e.g., learning rates, maximum training steps), following the format used in Caffe Solver Prototxt~\cite{caffeSolver}. 
    \item The objectives of the CNN pruning, including the  constraints on model size or accuracy.
\end{itemize}

The body of the CoCo-Tune framework consists of four main components as shown in Figure~\ref{fig:overview}. 
(1) The {\em hierarchical tuning block identifier} tries to define the set of {\em tuning blocks}. A {\bf tuning block} is a sequence of pruned consecutive CNN layers taken as a unit for pre-training. Suitable definitions of {\em tuning blocks} help maximize reuse while minimizing the pre-training overhead. (2) From the given CNN model specified in Prototxt, the {\em CoCo-Tune compiler} generates a {\em multiplexing model}, which is a function written in TensorFlow that, when invoked, specifies the structure of the full to-be-pruned CNN model, the network structure---which implements a Teacher-Student scheme---for pre-training tuning blocks, or pruned networks assembled with pre-trained tuning blocks, depending on the arguments the function receives. (3) The {\em pre-training scripts} are some generic Python functions that, when run, pre-train each tuning block based on the outputs from the first two components of CoCo-Tune. (4) The final component, {\em exploration scripts}, explores the promising pruned networks assembled with the pre-trained tuning blocks. The exploration of a network includes first fine-tuning the entire network and then testing it for accuracy. The exploration order is automatically picked by the {\em exploration scripts} based on the pruning objectives to produce the best network as early as possible. Both the {\em pre-training scripts} and the {\em exploration scripts} can run on one machine or multiple machines in a distributed environment through MPI.

The illustration is for a scenario where the promising subspace is given. For scenarios where the promising subspace is fully known at the beginning (e.g., the ADMM-based pruning mentioned in Section~\ref{sec:cocogen}), CoCo-Tune helps in shortening the tuning process of the appropriate pruning rate for each layer of the neural network. We next give a deeper view of each of the main components of CoCo-Tune. 

\paragraph{Hierarchical Compression-Based Algorithm}

Composability-based CNN pruning faces a trade-off between the pre-training cost and the time savings the pre-training results bring. The tradeoff depends on the definitions of the unit for pre-training, that is, the definition of {\em tuning blocks}. A {\em tuning block} is a unit for pre-training; it consists of a sequence of consecutive CNN layers pruned at certain rates. It can have various sizes, depending on the number of CNN layers it contains. The smaller it is, the less pre-training time it takes and the more reuses it tends to have across networks, but at the same time, its impact to the training time of a network tends to be smaller.

\begin{figure}
\centering
\includegraphics[width=.7\columnwidth]{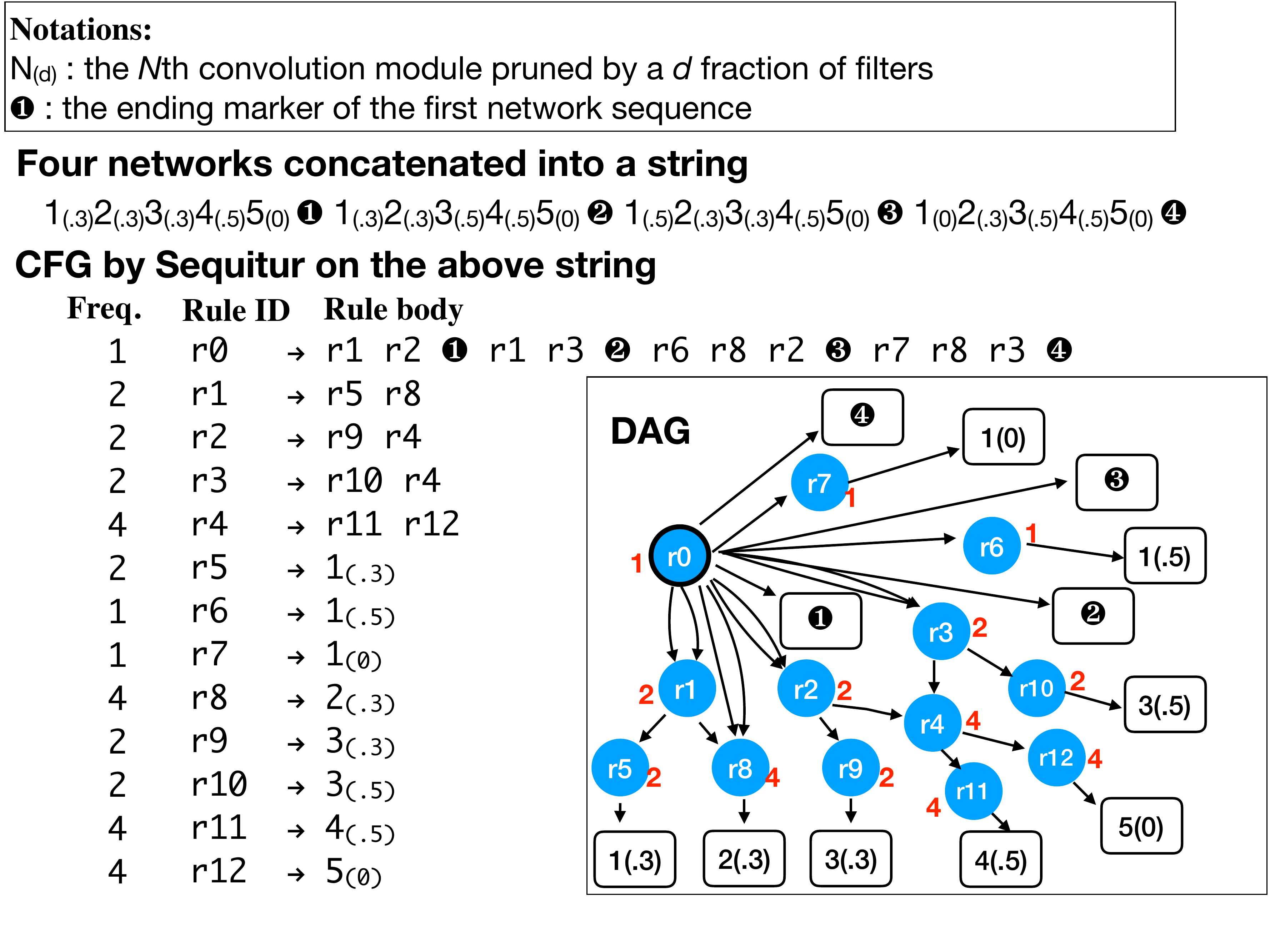}
\caption{Sequitur applies to a concatenated sequence of layers of four networks pruned at rates: 0\%, 30\%, 50\%.}
\label{fig:sequitur}
\end{figure}

To strike a good tradeoff between the pre-training cost and the benefits, we propose a hierarchical compression-baesd algorithm to help identify the best set of tuning blocks. 
Our algorithm leverages Sequitur~\cite{nevill1997identifying} to efficiently identify the frequent sequences of pruned layers in the network collection $C$. As a linear-time hierarchical compression algorithm, Sequitur infers a hierarchical
structure from a sequence of discrete symbols. For a given sequence of
symbols, it derives a context-free grammar (CFG), with each rule in
the CFG reducing a repeatedly appearing string into a single rule
ID. Figure~\ref{fig:sequitur} gives an example. Its top part shows the concatenated sequence of layers of four networks pruned at various rates; the subscripts of the numbers indicate the pruning rate, that is, the fraction of the least important filters of a layer that are removed. The lower part in Figure~\ref{fig:sequitur} shows the CFG produced by Sequitur on the string. A full expansion of rule $r0$ would give the original string. The result can also be represented as a Directed Acyclic Graph (DAG) as the right graph in Figure~\ref{fig:sequitur} shows with each node corresponding to one rule.

Applying Sequitur to the concatenated sequence of all networks in the {\em promising subspace}, our {\em hierarchical compression-based algorithm} gets the corresponding CFG and the DAG. Let $R$ be the collection of all the rules in the CFG, and $S$ be the solution to the tuning block identification problem which is initially empty. Our algorithm then heuristically fills $S$ with subsequences of CNN layers (represented as rules in the CFG) that are worth pre-training. 

It does it based on the appearing frequencies of the rules in the {\em promising subspace} and their sizes (i.e., the number of layers a rule contains). It employs two heuristics: (1) A rule cannot be put into $S$ if it appears in only one network (i.e., its appearing frequency is one); (2) a rule is preferred over its children rules only if that rule appears as often as its most frequently appearing descendant. 

\paragraph{Pre-Training of Tuning Blocks}


The standard CNN back propagation training algorithm uses input labels as the ground truth to compute errors of the current network and adjusts the weights iteratively. To train a tuning block, the first question is what ground truth to use to compute errors. Inspired by Teacher-Student networks~\cite{bucilua2006model,ba2014deep,hinton2015distilling}, we adopt a similar Teacher-Student mechanism to address the problem. 

\begin{figure}
  \centering
  \includegraphics[width=.7\columnwidth]{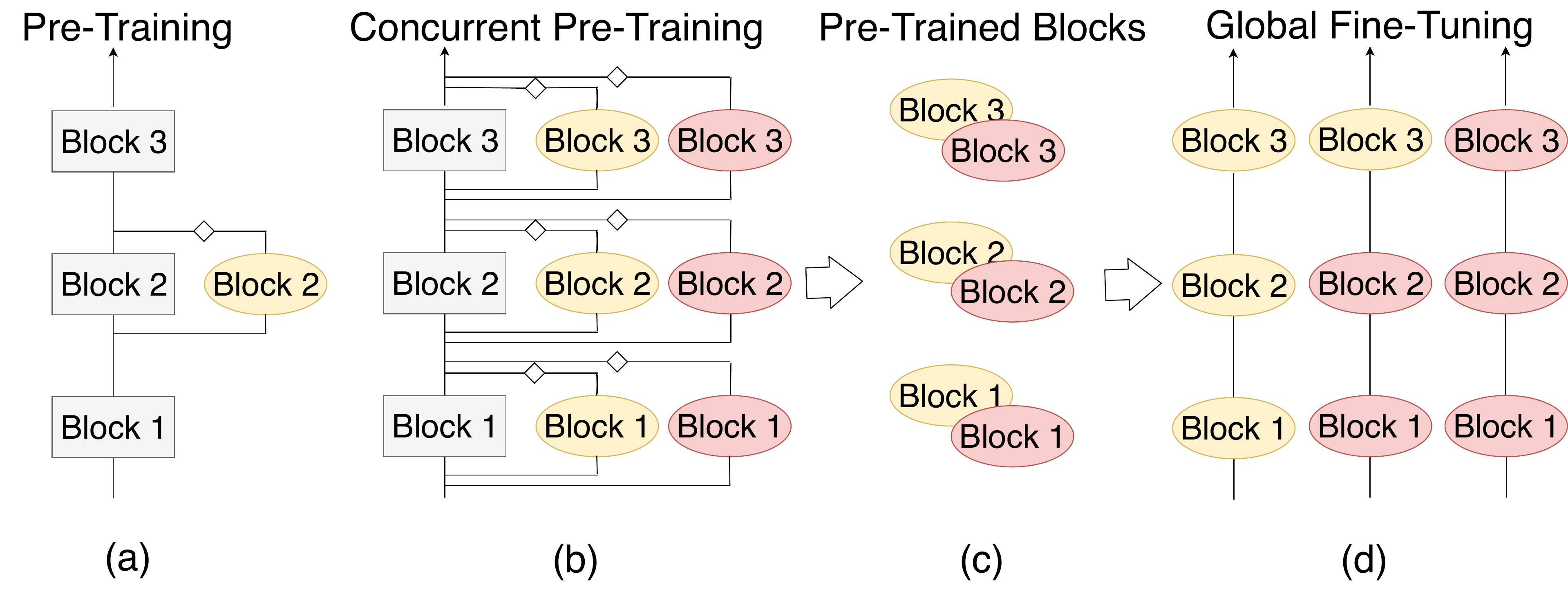}
  \caption{Illustration of composability-based network pruning. Eclipses are pruned tuning blocks; rectangles are original tuning blocks; diamonds refer to the activation map reconstruction error. Different colors of pruned tuning blocks correspond to different pruning options.
    }
  \label{fig:localtrain}
\end{figure}

We construct a network structure that contains both the pruned block to pre-train and the original full CNN model. They are put side by side as shown in Figure~\ref{fig:localtrain} (a) with the input to the counterpart of the tuning block in the full model also flowing into the pruned tuning block as its input, and the output activation map of the counterpart block flowing into the pruned tuning block as the "ground truth" of its output. When the standard back propagation algorithm is applied to the tuning block in this network structure, it effectively minimizes the reconstruction error between the output activation maps from the pruned tuning block and the ones from its unpruned counterpart in the full network. (In CNN pruning, the full model has typically already been trained beforehand to perform well on the datasets of interest.) This design essentially uses the full model as the "teacher" to train the pruned tuning blocks. 

This Teacher-Student design has three appealing properties. First, it addresses the missing ``ground truth" problem for tuning block pre-training. Second, as the full CNN model runs along with the pre-training of the tuning blocks, it provides the inputs and "ground truth" for the tuning blocks on the fly; there is no need to save to storage the activation maps which can be space-consuming considering the large number of input images for training a CNN. Third, the structure is friendly for concurrently pre-training multiple tuning blocks. As Figure~\ref{fig:localtrain} (b) shows, connections can be added between the full model and multiple pruned blocks; the pre-training of these blocks can then happen in one run, and the activation maps produced by a block in the full model can be seamlessly reused across the pre-training of multiple pruned blocks. 

\paragraph{Global Fine-Tuning}

The local training phase outputs a bag of pre-trained pruned tuning blocks, as shown in Figure~\ref{fig:localtrain} (c) (tuning blocks in the original network could also be included). At the beginning of the {\em global fine-tuning} phase is an assembly step, which, logically, assembles these training blocks into each of the networks in the promising subspace. Physically, this step just needs to initialize the pruned networks in the promising subspace with the weights in the corresponding tuning blocks. We call the resulting network {\em a block-trained network}. Recall that one of the side products of the tuning block identification step is a {\em composite vector} for each network which records the tuning blocks the network can use; these vectors are used in this assembly step. Figure~\ref{fig:localtrain} (d) gives a conceptual illustration; three networks are assembled with different sets of pre-trained tuning blocks.


As a pruned block with only a subset of parameters has a smaller model capacity, a {\em global fine-tuning step} is required to further recover the accuracy performance of a block-trained network. This step runs the standard CNN training on the {\em block-trained networks}. All the parameters in the networks are updated during the training.
Compared with training a default pruned network, fine-tuning a block-trained network usually takes much less training time as the network starts with a much better set of parameter values as shown later in this article. 



\paragraph{CoCo-Tune Compiler and Scripts}

CoCo-Tune compiler and scripts offer an automatic way to materialize the mechanisms described in the earlier parts of this section for an arbitrary CNN model. The proposed method is not restricted to a particular DNN framework, though we demonstrate its ability using TensorFlow. 

TensorFlow APIs with other assistant libraries (e.g., Slim~\cite{slim}) offer conveniences for standard CNN model training and testing, but {\em not} for CNN pruning, let alone {\em composability-based pruning}. Asking a general programmer to implement {\em composability-based pruning} in TensorFlow for each CNN model would add tremendous burdens on the programmer. She would need to write code to identify tuning blocks, create TensorFlow code to implement the customized CNN structures to pre-train each tuning block, generate checkpoints, and use them when creating the block-trained CNN networks for global fine-tuning.

CoCo-Tune compiler and scripts mitigate the difficulty by automating the process. The fundamental motivating observation is that the codes for two different CNN models follow the same pattern. Differences are mostly on the code specifying the structure of the CNN models (both the original and the extended for pre-training and global fine tuning). The idea is to build code templates and use the compiler to automatically adapt the templates based on the specifications of the models. 







CoCo-Tune takes Prototxt as the format of an input to-be-pruned CNN model. 
It first generates a \textit{multiplexing model}, which is a piece of TensorFlow code defined as a Python function. It is multiplexing in the sense that an invocation of the code specifies the structure of the original CNN model, or the structure for pre-training, or the global fine tuning model; which of the three modes is used at an invocation of the multiplexing model is determined by one of its input arguments, {\em mode\_to\_use}. The multiplexing design allows easy code reuse as the three modes share much common code for model specifications. Another argument, {\em prune\_info}, conveys to the multiplexing model the pruning information, including the set of tuning blocks to pre-train in this invocation and their pruning rates. The compiler generates code that maps CNN model specifications in Prototxt to TensorFlow APIs, specifies the derived network structure for pre-training each tuning block contained in {\em prune\_info}, and conducts global fine-tuning of the pruned networks assembled from the pre-trained tuning blocks.

The compiler automates the composability-based pruning, making CoCo-Tune a tool for both speed and productivity for DNN pruning.

\subsubsection{Evaluations}
We conduct a set of experiments to examine the efficacy of CoCo-Tune. Our experiments use four popular CNN models: ResNet-50 and ResNet-101, as representatives of the Residual Network family~\cite{he2016deep}, and Inception-V2 and Inception-V3, as representatives of the Inception family~\cite{szegedy2015going}. They have 50, 101, 34, 48 layers respectively. These models represent a structural trend in CNN designs, in which, several layers are encapsulated into a generic module of a fixed structure---which we call {\em convolution module}---and a network is built by stacking many such modules together. 
Such CNN models are holding the state-of-the-art accuracy in many challenging deep learning tasks. The structures of these models are described in input Caffe Prototxt\footnote{We add to Prototxt a new construct "module" for specifying the boundaries of {\em convolution modules}.} files and converted to the multiplexing models by the CoCo-Tune compiler. 

For preparation, we adapt the four CNN models trained on a general image dataset ImageNet~\cite{russakovsky2015imagenet} (ILSVRC 2012) to each of four specific image classification tasks with the domain-specific datasets, Flowers102~\cite{nilsback2008automated}, CUB200~\cite{welinder2010caltech}, Cars~\cite{krause20133d}, and Dogs~\cite{khosla2011novel}. It gives us 16 trained full CNN models. The accuracy of the trained ResNets and Inceptions  on the test datasets are listed in columns \textit{Accuracy} in Table~\ref{tab:dataset}. 
The four datasets for CNN pruning are commonly used in fine-grained recognition~\cite{krause2016unreasonable,fu2017look,molchanov2016pruning,howard2017mobilenets,zhao2017diversified}, which is a typical usage scenario of CNN pruning. Table \ref{tab:dataset} reports the statistics of the four datasets, including the data size for training (\textit{Train}), the data size for testing (\textit{Test}), and the number of classes (\textit{Classes}). For all experiments, network training is performed on the training sets while accuracy results are reported on the testing sets.

\begin{table*}
  \centering
  \caption{Dataset statistics.}
  \label{tab:dataset}
\scriptsize
\vspace*{-0.5\baselineskip}
\begin{tabular}{|l|l|l|l|l|l|l|l|l|l|}
  \hline
   \multirow{2}{*}{} & \multirow{2}{*}{Dataset} & \multicolumn{3}{c|}{Size} & \multirow{2}{*}{Classes} & \multicolumn{4}{c|}{Accuracy} \\ \cline{3-5} \cline{7-10} 
  & & Total & Train & Test &  & ResNet-50 & ResNet-101 & Inception-V2 & Inception-V3  \\ \hline \hline
  General &  ImageNet~\cite{russakovsky2015imagenet} & 1,250,000 & 1,200,000 & 50,000  & 1000 & 0.752 & 0.764 & 0.739 & 0.780 \\ \hline \hline
  \multirow{4}{*}{Special} &  Flowers102~\cite{nilsback2008automated} &  8,189 & 6,149 & 2,040 & 102 & 0.973 & 0.975 & 0.972 & 0.968 \\ \cline{2-10}
  & CUB200~\cite{welinder2010caltech} &  11,788 & 5,994 & 5,794 & 200 & 0.770 & 0.789 & 0.746 & 0.760 \\ \cline{2-10}
  & Cars~\cite{krause20133d} & 16,185 & 8,144 & 8,041 & 196 & 0.822 & 0.845 & 0.789 & 0.801 \\ \cline{2-10}
  & Dogs~\cite{khosla2011novel}&  20,580 & 12,000 & 8,580 & 120 & 0.850 & 0.864 & 0.841 & 0.835 \\ \hline
\end{tabular}
\end{table*}

\paragraph{Baseline for Comparison}
In CNN pruning, the full CNN model to prune has typically been already trained on the datasets of interest. When filters in the CNN are pruned, a new model with fewer filters is created, which inherits the remaining parameters of the affected layers and the unaffected layers in the full model. The promising subspace consists of such models. The {\em baseline approach} trains these models as they are. \blue{Although there are prior studies on accelerating CNN pruning, what they propose are all various ways to reduce the configuration space to a promising subspace.} To the best of our knowledge, when exploring the configurations in the promising subspace, they all use the {\em baseline approach}. As our method is the first for speeding up the exploration of the promising space, we compare our results with those from the {\em baseline approach}. 

We refer to a pruned network in the baseline approach a \textit{default network} while the one initialized with pre-trained tuning blocks in our method a \textit{block-trained network}. 

\paragraph{Promising Subspace} 
The 16 trained CNNs contain up to hundreds of convolutional layers. A typical practice is to use the same pruning rate for the convolutional layers in one {\em convolution module}. We adopt the same strategy. 
The importance of a filter is determined by its $\ell_1$ norm as previous work~\cite{li2016pruning} proposes. 
Following prior CNN pruning practice~\cite{li2016pruning,luo2017thinet}, the top layer of a convolution module is kept unpruned; it helps ensure the dimension compatibility of the module. 


There are many ways to select the promising subspace, i.e., the set of promising configurations worth evaluating. Previous works select configurations either manually~\cite{li2016pruning,luo2017thinet} or based on reinforcement learning with various rewards or algorithm design~\cite{he2018adc,ashok2017n2n}. As that is orthogonal to the focus of this work, to avoid bias from that factor, our experiment forms the promising spaces through random sampling~\cite{bergstra2012random} of the entire pruning space. A promising space contains 500 pruned networks, whose sizes follow a close-to-uniform distribution. In the experiments, the pruning rate for a layer can be one of $\Gamma=\{30\%, 50\%, 70\%\}$. 

\paragraph{Objective of Pruning}
There are different pruning objectives including minimizing model size, computational cost, memory footprint or energy consumption. 
Even though an objective of pruning affects the choice of the best configuration, all objectives require the evaluation of the set of promising configurations. Our composability-based CNN pruning aims at accelerating the training of a set of pruned networks and thus can work with any objective of pruning. 

For the demonstration purpose, we set the objective of pruning as finding the smallest network ({min ModelSize}) that meets a given accuracy threshold ({Accuracy $<=$ \textit{thr\_acc}}). We get a spectrum of \textit{thr\_acc} values by varying the accuracy drop rate $\alpha$ from that of the full model from -0.02 to 0.08. We include negative drop rates because it is possible that pruning makes the model more accurate.

All the experiments are performed with TensorFlow 1.3.0 on machines each equipped with a 16-core 2.2GHz AMD Opteron 6274 (Interlagos) processor, 32 GB of RAM and an NVIDIA K20X GPU with 6 GB of DDR5 memory. One network is trained on one GPU.

To measure the basic benefits from the composability-based method, these experiments use every convolution module in these networks as a tuning block. The extra benefits from hierarchical tuning block identification are reported later.


\begin{figure}
\centering
  \subfloat[Flowers102\label{fig:flowers102_resnet_accuracy_with_model_size}]{%
    \includegraphics[width=0.25\textwidth]{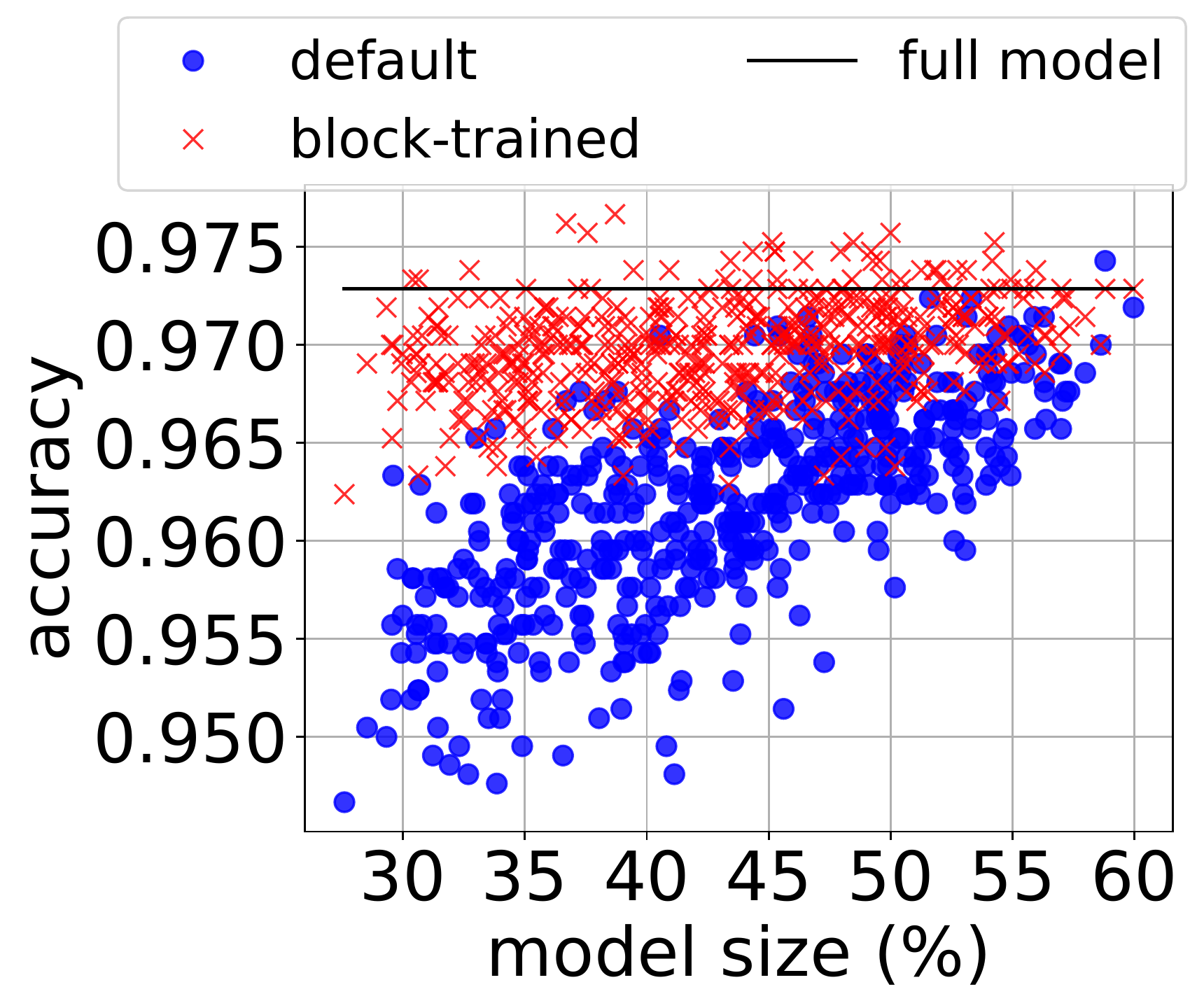} 
  } 
  \subfloat[Cars\label{fig:cars_resnet_accuracy_with_model_size}]{%
    \includegraphics[width=0.25\textwidth]{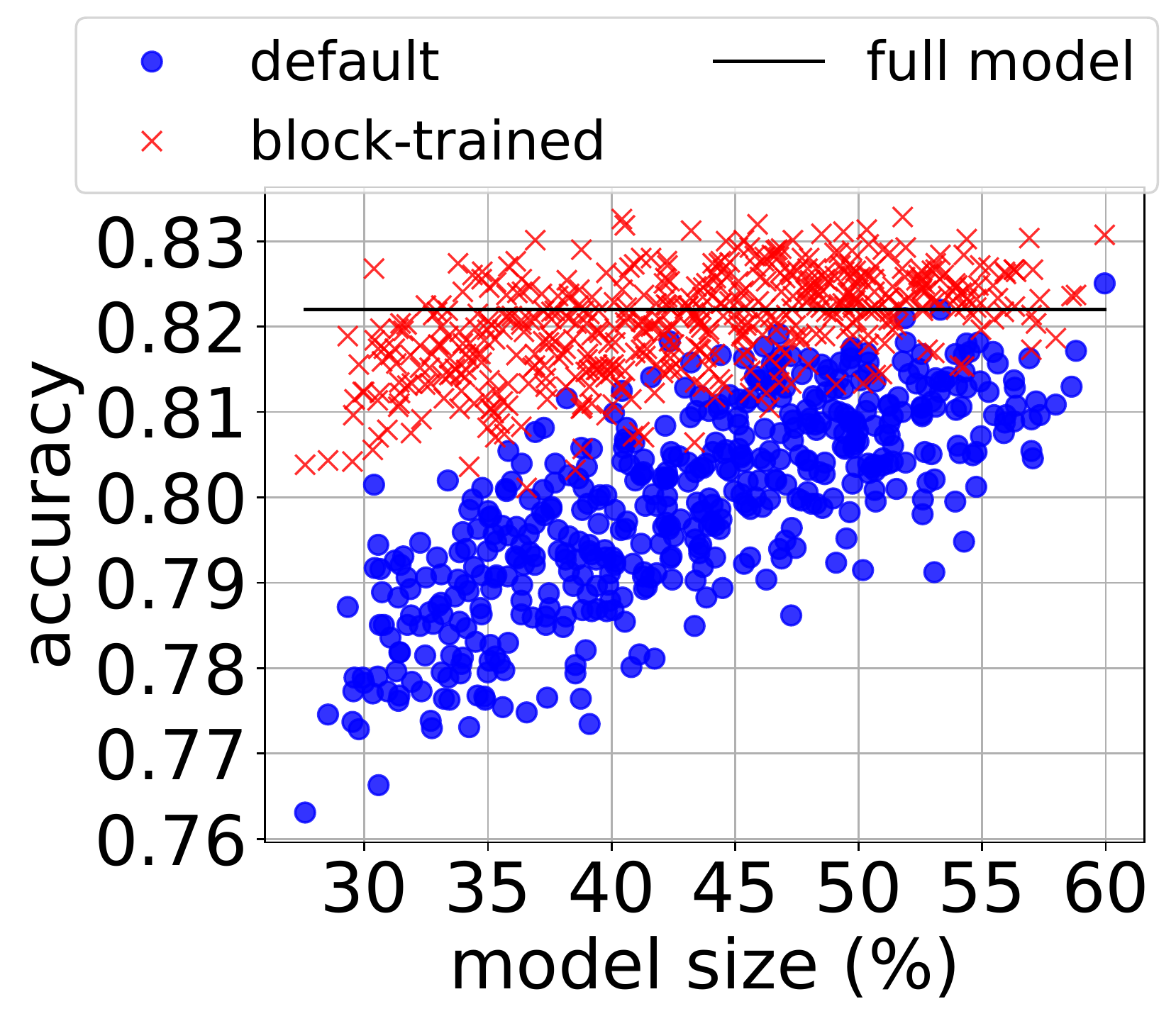} 
  }
  \subfloat[ResNet-50\label{fig:cub200_resnet_id0_annotated}]{%
    \includegraphics[width=0.25\textwidth]{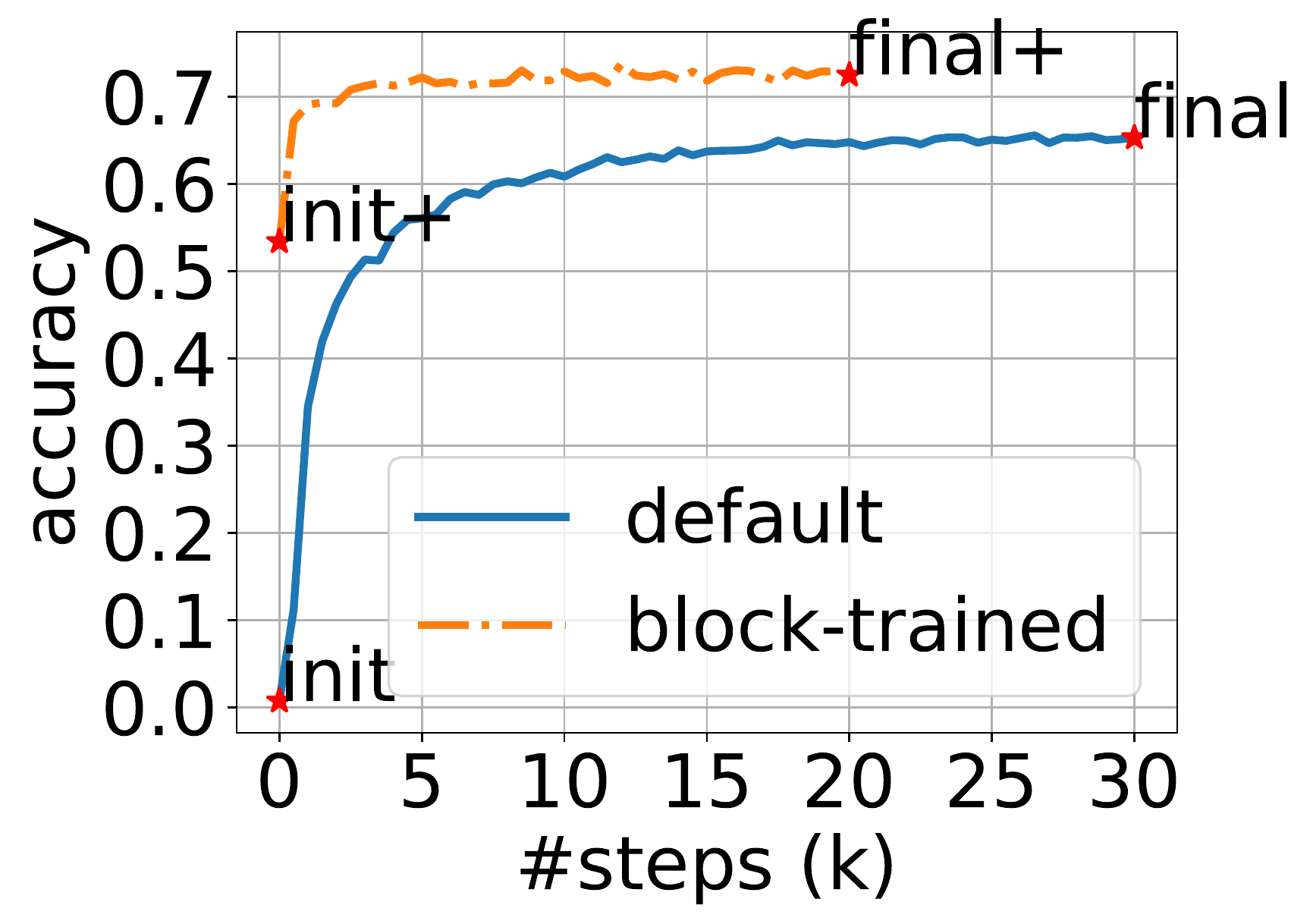} 
  } 
  \subfloat[Inception-V3\label{fig:2}]{%
    \includegraphics[width=0.25\textwidth]{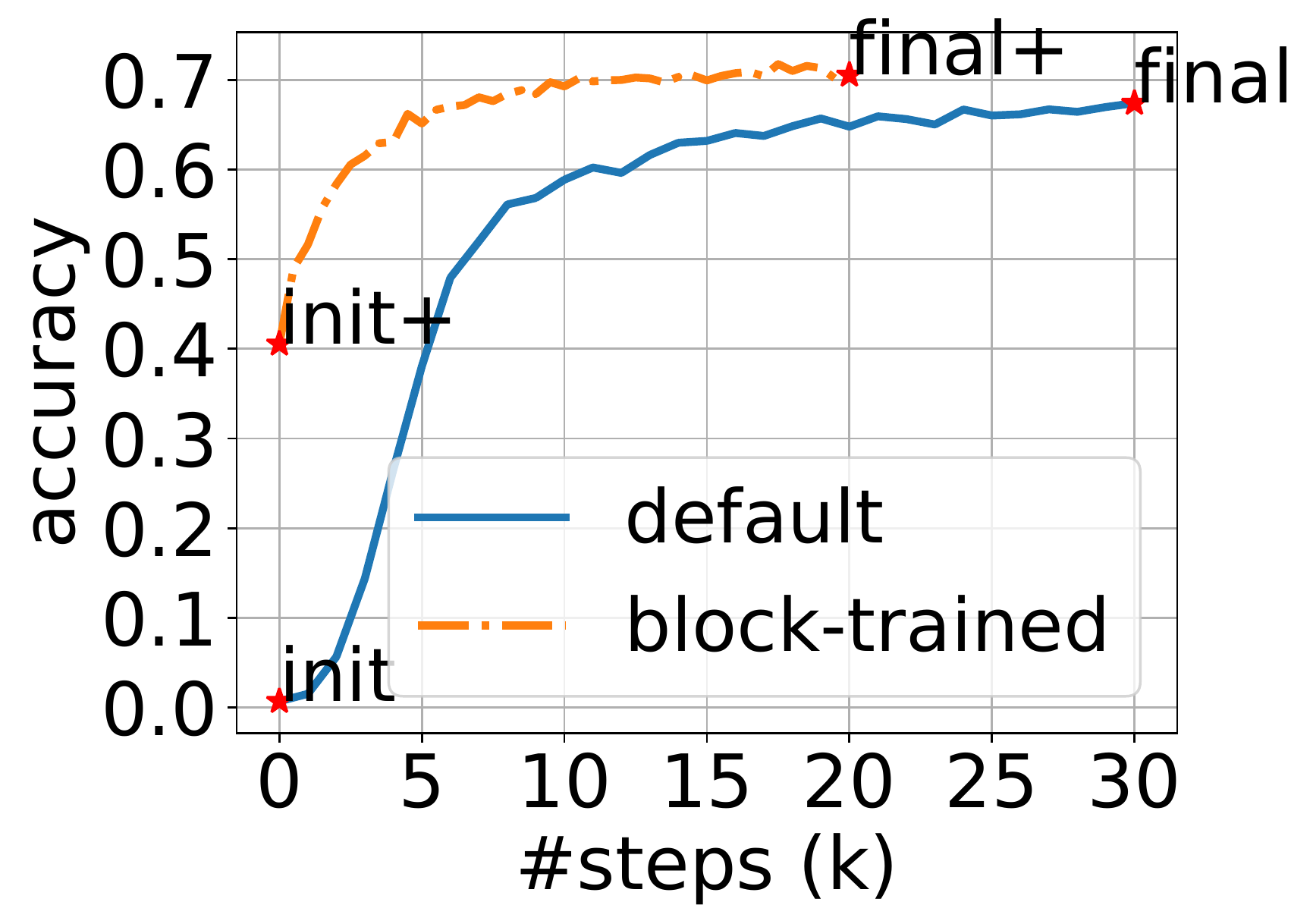} 
  } 
  \caption{(a,b) Accuracies of pruned networks of ResNet-50 after training. The model size of full ResNet-50 is 25.6 million. (c,d) Accuracy curves of the \textit{default} and \textit{block-trained} networks on dataset CUB200; each network has 70\% least important filters pruned at all convolution modules.}
  \label{fig:accuracy_with_model_size}
  	\vspace*{-1\baselineskip}
\end{figure}

Figure \ref{fig:accuracy_with_model_size} (a,b) show the final accuracies of all the 500 ResNet-50 variants trained with or without leveraging composability on the Flower102 and CUB200 datasets. For reference, we also plot the accuracies of the well-trained full ResNet-50 on the two datasets. The block-trained network gives a clearly better final accuracy overall.  

Table~\ref{tab:seq_speedups} reports the comparisons between the block-trained version and the default version, in both speeds and network sizes, at various levels of tolerable accuracy drop rates $\alpha$ (negative means higher accuracy than the large network gives). The results are collected when 1, 4, or 16 machines are used for concurrent training for both the baseline and our method (indicated by the "\#nodes" column). The time of the block-trained version already takes the pre-training time of tuning blocks into account ("overhead" in Table~\ref{tab:seq_speedups} shows the percentage in overall time). For the objective of pruning, the exploration order CoCo-Tune adopts is to start from the smallest models and proceed to larger ones. 

The results show that the composability-based method avoids up to $99.6\%$ of trial configurations and reduces the evaluation time by up to 186X for  ResNet-50; up to $96.7\%$ reduction and 30X speedups for Inception-V3. The reduction of trial configurations is because the method improves the accuracy of the pruned networks as Figure~\ref{fig:accuracy_with_model_size} shows. As a result, the exploration meets a desirable configuration sooner. For instance, in Flower102 ($\alpha=0$), the third smallest network can already reach the target accuracy in the block-trained version while the 297th network meets the target in the default version. This not only shortens the exploration time, but also yields more compact (up to 70\% smaller) networks as the ``model size'' columns in Table~\ref{tab:seq_speedups} show. Another reason for the speedup is that the training of a block-trained network takes fewer iterations to reach its final accuracy level than the default version, as Figure~\ref{fig:accuracy_with_model_size} (c,d) show. Even when configurations are not reduced (e.g., Flower102, $\alpha=-1$), the block-trained exploration finishes sooner. 

\blue{Table~\ref{tab:different_subspace_size} shows the speedups by composability-based pruning with different subspace sizes. The speedups are higher as the number of configurations to explore increases.  It is because the time for pre-training tuning blocks weights less as the total time increases and the reduction of configurations becomes more significant for a larger set. Another observation is that, when the number of configurations is only four, there is still a significant speedup in most cases. The block training time is the time spent on pre-training all the tuning block variants (48 for ResNet-50 and 27 for Inception-V3). The speedup could be higher if tuning block identifier is applied, as shown next.}
\begin{table*}
\tiny
  \centering 
  \tabcolsep=0.06cm
  \caption{Speedups and configuration savings by composability-based pruning (when 1, 4, or 16 machines are used for both baseline and composability-based methods as "\#nodes" column indicates). Notations are at the table bottom.}
  \vspace{-.5\baselineskip}
  \label{tab:seq_speedups}
\begin{tabular}{|l|l|l|l|l|l|l|l|l|l|l|l||l|l|l|ll|l|l|l|l|}
\hline
\multirow{3}{*}{Dataset} & \multirow{3}{*}{$\alpha$} & \multirow{3}{*}{\#nodes} & \multicolumn{9}{c||}{ResNet-50} & \multicolumn{9}{c|}{Inception-V3} \\ \cline{4-21} 
 &  &  & \multirow{2}{*}{thr\_acc} & \multicolumn{2}{c|}{\#configs} & \multicolumn{2}{c|}{\blue{time (h)}} & \multicolumn{2}{c|}{model size} & \multirow{2}{*}{\begin{tabular}[c]{@{}l@{}}speedup\\ (X)\end{tabular}} & \multirow{2}{*}{overhead} & \multirow{2}{*}{thr\_acc} & \multicolumn{2}{c|}{\#configs} & \multicolumn{2}{c}{\blue{time (h)}} & \multicolumn{1}{c|}{model size} &  & \multirow{2}{*}{\begin{tabular}[c]{@{}l@{}}speedup \\ (X)\end{tabular}} & \multirow{2}{*}{overhead} \\ \cline{5-10} \cline{14-19}
 &  &  &  & base & comp & base & comp & base & comp &  &  &  & base & comp & \multicolumn{1}{l|}{base} & comp & base & comp &  &  \\ \hline\hline
\multirow{3}{*}{Flowers102} & -1\% & \begin{tabular}[c]{@{}l@{}}1\\ 4\\ 16\end{tabular} & 0.983 & \begin{tabular}[c]{@{}l@{}}500\\ 500\\ 500\end{tabular} & \begin{tabular}[c]{@{}l@{}}500\\ 500\\ 500\end{tabular} & \begin{tabular}[c]{@{}l@{}}2858.7\\ 718.1\\ 184.9\end{tabular} & \begin{tabular}[c]{@{}l@{}}1912.7\\ 481.0\\ 125.5\end{tabular} & 100\% & 100\% & \begin{tabular}[c]{@{}l@{}}1.5\\ 1.5\\ 1.5\end{tabular} & \begin{tabular}[c]{@{}l@{}}0.4\%\\ 0.5\%\\ 1.8\%\end{tabular} & 0.978 & \begin{tabular}[c]{@{}l@{}}500\\ 500\\ 500\end{tabular} & \begin{tabular}[c]{@{}l@{}}500\\ 500\\ 500\end{tabular} & \multicolumn{1}{l|}{\begin{tabular}[c]{@{}l@{}}3018.8\\ 756.7\\ 194.8\end{tabular}} & \begin{tabular}[c]{@{}l@{}}2023.5\\ 508.1\\ 133.6\end{tabular} & 100\% & 100\% & \begin{tabular}[c]{@{}l@{}}1.5\\ 1.5\\ 1.5\end{tabular} & \begin{tabular}[c]{@{}l@{}}0.5\%\\ 0.7\%\\ 2.7\%\end{tabular} \\ \cline{2-21} 
 & 0\% & \begin{tabular}[c]{@{}l@{}}1\\ 4\\ 16\end{tabular} & 0.973 & \begin{tabular}[c]{@{}l@{}}297\\ 300\\ 304\end{tabular} & \begin{tabular}[c]{@{}l@{}}3\\ 4\\ 16\end{tabular} & \begin{tabular}[c]{@{}l@{}}1639.4\\ 412.6\\ 103.3\end{tabular} & \begin{tabular}[c]{@{}l@{}}16.9\\ 5.2\\ 4.7\end{tabular} & 45.4\% & 29.3\% & \begin{tabular}[c]{@{}l@{}}97.0\\ 79.3\\ 22.0\end{tabular} & \begin{tabular}[c]{@{}l@{}}40.4\%\\ 43.5\%\\ 48.3\%\end{tabular} & 0.968 & \begin{tabular}[c]{@{}l@{}}244\\ 244\\ 256\end{tabular} & \begin{tabular}[c]{@{}l@{}}10\\ 12\\ 16\end{tabular} & \multicolumn{1}{l|}{\begin{tabular}[c]{@{}l@{}}1428.6\\ 358.2\\ 94.8\end{tabular}} & \begin{tabular}[c]{@{}l@{}}47.3\\ 13.9\\ 6.5\end{tabular} & 43.2\% & 32.4\% & \begin{tabular}[c]{@{}l@{}}30.2\\ 25.8\\ 14.6\end{tabular} & \begin{tabular}[c]{@{}l@{}}23.3\%\\ 26.4\%\\ 56.4\%\end{tabular} \\ \cline{2-21} 
 & 1\% & \begin{tabular}[c]{@{}l@{}}1\\ 4\\ 16\end{tabular} & 0.963 & \begin{tabular}[c]{@{}l@{}}6\\ 8\\ 16\end{tabular} & \begin{tabular}[c]{@{}l@{}}1\\ 4\\ 16\end{tabular} & \begin{tabular}[c]{@{}l@{}}31.0\\ 10.4\\ 5.2\end{tabular} & \begin{tabular}[c]{@{}l@{}}8.3\\ 3.2\\ 2.9\end{tabular} & 29.6\% & 27.6\% & \begin{tabular}[c]{@{}l@{}}3.7\\ 3.3\\ 1.8\end{tabular} & \begin{tabular}[c]{@{}l@{}}82.8\%\\ 70.6\%\\ 78.3\%\end{tabular} & 0.958 & \begin{tabular}[c]{@{}l@{}}27\\ 28\\ 32\end{tabular} & \begin{tabular}[c]{@{}l@{}}1\\ 4\\ 16\end{tabular} & \multicolumn{1}{l|}{\begin{tabular}[c]{@{}l@{}}152.6\\ 39.6\\ 11.2\end{tabular}} & \begin{tabular}[c]{@{}l@{}}13.9\\ 5.8\\ 5.6\end{tabular} & 33.9\% & 31.0\% & \begin{tabular}[c]{@{}l@{}}11.0\\ 6.8\\ 2.2\end{tabular} & \begin{tabular}[c]{@{}l@{}}79.0\%\\ 63.3\%\\ 71.0\%\end{tabular} \\ \hline\hline
\multirow{3}{*}{CUB200} & 4\% & \begin{tabular}[c]{@{}l@{}}1\\ 4\\ 16\end{tabular} & 0.739 & \begin{tabular}[c]{@{}l@{}}323\\ 324\\ 336\end{tabular} & \begin{tabular}[c]{@{}l@{}}2\\ 4\\ 16\end{tabular} & \begin{tabular}[c]{@{}l@{}}1807.3\\ 454.0\\ 118.7\end{tabular} & \begin{tabular}[c]{@{}l@{}}12.7\\ 3.1\\ 3.1\end{tabular} & 46.6\% & 28.5\% & \begin{tabular}[c]{@{}l@{}}142.3\\ 146.5\\ 38.3\end{tabular} & \begin{tabular}[c]{@{}l@{}}53.7\%\\ 74.4\%\\ 74.4\%\end{tabular} & 0.720 & \begin{tabular}[c]{@{}l@{}}74\\ 76\\ 80\end{tabular} & \begin{tabular}[c]{@{}l@{}}3\\ 4\\ 16\end{tabular} & \multicolumn{1}{l|}{\begin{tabular}[c]{@{}l@{}}420.2\\ 106.4\\ 27.6\end{tabular}} & \begin{tabular}[c]{@{}l@{}}21.9\\ 6.7\\ 6.0\end{tabular} & 41.4\% & 33.7\% & \begin{tabular}[c]{@{}l@{}}19.2\\ 15.9\\ 4.6\end{tabular} & \begin{tabular}[c]{@{}l@{}}49.8\%\\ 54.5\%\\ 60.6\%\end{tabular} \\ \cline{2-21} 
 & 5\% & \begin{tabular}[c]{@{}l@{}}1\\ 4\\ 16\end{tabular} & 0.731 & \begin{tabular}[c]{@{}l@{}}297\\ 300\\ 304\end{tabular} & \begin{tabular}[c]{@{}l@{}}1\\ 4\\ 16\end{tabular} & \begin{tabular}[c]{@{}l@{}}1654.7\\ 418.8\\ 105.5\end{tabular} & \begin{tabular}[c]{@{}l@{}}8.9\\ 2.8\\ 2.7\end{tabular} & 45.4\% & 27.6\% & \begin{tabular}[c]{@{}l@{}}185.9\\ 149.6\\ 39.1\end{tabular} & \begin{tabular}[c]{@{}l@{}}77.1\%\\ 81.4\%\\ 83.7\%\end{tabular} & 0.710 & \begin{tabular}[c]{@{}l@{}}44\\ 44\\ 48\end{tabular} & \begin{tabular}[c]{@{}l@{}}1\\ 4\\ 16\end{tabular} & \multicolumn{1}{l|}{\begin{tabular}[c]{@{}l@{}}247.8\\ 61.7\\ 16.4\end{tabular}} & \begin{tabular}[c]{@{}l@{}}14.1\\ 5.4\\ 5.2\end{tabular} & 38.5\% & 31.5\% & \begin{tabular}[c]{@{}l@{}}17.6\\ 11.4\\ 3.2\end{tabular} & \begin{tabular}[c]{@{}l@{}}77.5\%\\ 67.6\%\\ 70.6\%\end{tabular} \\ \cline{2-21} 
 & 6\% & \begin{tabular}[c]{@{}l@{}}1\\ 4\\ 16\end{tabular} & 0.724 & \begin{tabular}[c]{@{}l@{}}154\\ 156\\ 160\end{tabular} & \begin{tabular}[c]{@{}l@{}}1\\ 4\\ 16\end{tabular} & \begin{tabular}[c]{@{}l@{}}840.1\\ 214.2\\ 53.8\end{tabular} & \begin{tabular}[c]{@{}l@{}}8.3\\ 2.6\\ 2.5\end{tabular} & 38.0\% & 27.6\% & \begin{tabular}[c]{@{}l@{}}101.2\\ 82.4\\ 21.5\end{tabular} & \begin{tabular}[c]{@{}l@{}}82.6\%\\ 86.7\%\\ 89.7\%\end{tabular} & 0.700 & \begin{tabular}[c]{@{}l@{}}29\\ 32\\ 32\end{tabular} & \begin{tabular}[c]{@{}l@{}}1\\ 4\\ 16\end{tabular} & \multicolumn{1}{l|}{\begin{tabular}[c]{@{}l@{}}162.5\\ 44.5\\ 10.8\end{tabular}} & \begin{tabular}[c]{@{}l@{}}12.8\\ 5.3\\ 5.1\end{tabular} & 35.9\% & 31.0\% & \begin{tabular}[c]{@{}l@{}}12.7\\ 8.4\\ 2.1\end{tabular} & \begin{tabular}[c]{@{}l@{}}85.1\%\\ 68.7\%\\ 71.9\%\end{tabular} \\ \hline\hline
\multirow{3}{*}{Cars} & -1\% & \begin{tabular}[c]{@{}l@{}}1\\ 4\\ 16\end{tabular} & 0.830 & \begin{tabular}[c]{@{}l@{}}500\\ 500\\ 500\end{tabular} & \begin{tabular}[c]{@{}l@{}}100\\ 100\\ 112\end{tabular} & \begin{tabular}[c]{@{}l@{}}2864.9\\ 720.4\\ 185.3\end{tabular} & \begin{tabular}[c]{@{}l@{}}362.4\\ 90.9\\ 27.1\end{tabular} & 100\% & 35.7\% & \begin{tabular}[c]{@{}l@{}}7.9\\ 7.9\\ 6.8\end{tabular} & \begin{tabular}[c]{@{}l@{}}1.9\%\\ 2.5\%\\ 8.4\%\end{tabular} & 0.811 & \begin{tabular}[c]{@{}l@{}}271\\ 272\\ 272\end{tabular} & \begin{tabular}[c]{@{}l@{}}20\\ 20\\ 32\end{tabular} & \multicolumn{1}{l|}{\begin{tabular}[c]{@{}l@{}}1586.8\\ 398.1\\ 99.4\end{tabular}} & \begin{tabular}[c]{@{}l@{}}85.6\\ 22.4\\ 11.1\end{tabular} & 40.1\% & 33.5\% & \begin{tabular}[c]{@{}l@{}}18.5\\ 17.8\\ 9.0\end{tabular} & \begin{tabular}[c]{@{}l@{}}12.8\%\\ 16.3\%\\ 32.8\%\end{tabular} \\ \cline{2-21} 
 & 0\% & \begin{tabular}[c]{@{}l@{}}1\\ 4\\ 16\end{tabular} & 0.822 & \begin{tabular}[c]{@{}l@{}}332\\ 332\\ 336\end{tabular} & \begin{tabular}[c]{@{}l@{}}11\\ 12\\ 16\end{tabular} & \begin{tabular}[c]{@{}l@{}}1848.6\\ 461.4\\ 115.9\end{tabular} & \begin{tabular}[c]{@{}l@{}}44.4\\ 12.1\\ 5.2\end{tabular} & 46.9\% & 30.4\% & \begin{tabular}[c]{@{}l@{}}41.6\\ 38.1\\ 22.3\end{tabular} & \begin{tabular}[c]{@{}l@{}}15.4\%\\ 18.8\%\\ 44.0\%\end{tabular} & 0.801 & \begin{tabular}[c]{@{}l@{}}84\\ 84\\ 96\end{tabular} & \begin{tabular}[c]{@{}l@{}}3\\ 4\\ 16\end{tabular} & \multicolumn{1}{l|}{\begin{tabular}[c]{@{}l@{}}480.3\\ 120.5\\ 33.8\end{tabular}} & \begin{tabular}[c]{@{}l@{}}21.8\\ 7.2\\ 6.7\end{tabular} & 36.9\% & 31.3\% & \begin{tabular}[c]{@{}l@{}}22.0\\ 16.7\\ 5.0\end{tabular} & \begin{tabular}[c]{@{}l@{}}50.2\%\\ 50.6\%\\ 54.7\%\end{tabular} \\ \cline{2-21} 
 & 1\% & \begin{tabular}[c]{@{}l@{}}1\\ 4\\ 16\end{tabular} & 0.814 & \begin{tabular}[c]{@{}l@{}}189\\ 192\\ 192\end{tabular} & \begin{tabular}[c]{@{}l@{}}2\\ 4\\ 6\end{tabular} & \begin{tabular}[c]{@{}l@{}}1026.4\\ 259.7\\ 65.5\end{tabular} & \begin{tabular}[c]{@{}l@{}}12.8\\ 4.9\\ 4.1\end{tabular} & 40.4\% & 28.5\% & \begin{tabular}[c]{@{}l@{}}80.2\\ 53.0\\ 16.0\end{tabular} & \begin{tabular}[c]{@{}l@{}}53.4\%\\ 46.7\%\\ 55.7\%\end{tabular} & 0.791 & \begin{tabular}[c]{@{}l@{}}33\\ 36\\ 48\end{tabular} & \begin{tabular}[c]{@{}l@{}}1\\ 4\\ 16\end{tabular} & \multicolumn{1}{l|}{\begin{tabular}[c]{@{}l@{}}186.4\\ 50.7\\ 16.4\end{tabular}} & \begin{tabular}[c]{@{}l@{}}14.2\\ 6.8\\ 6.2\end{tabular} & 34.4\% & 31.0\% & \begin{tabular}[c]{@{}l@{}}13.1\\ 7.5\\ 2.6\end{tabular} & \begin{tabular}[c]{@{}l@{}}77.0\%\\ 54.0\%\\ 59.1\%\end{tabular} \\ \hline\hline
\multirow{3}{*}{Dogs} & 6\% & \begin{tabular}[c]{@{}l@{}}1\\ 4\\ 16\end{tabular} & 0.799 & \begin{tabular}[c]{@{}l@{}}500\\ 500\\ 500\end{tabular} & \begin{tabular}[c]{@{}l@{}}123\\ 124\\ 128\end{tabular} & \begin{tabular}[c]{@{}l@{}}2848.1\\ 709.8\\ 178.0\end{tabular} & \begin{tabular}[c]{@{}l@{}}441.1\\ 111.2\\ 28.3\end{tabular} & 60.0\% & 36.9\% & \begin{tabular}[c]{@{}l@{}}6.5\\ 6.4\\ 6.3\end{tabular} & \begin{tabular}[c]{@{}l@{}}1.6\%\\ 2.0\%\\ 8.1\%\end{tabular} & 0.776 & \begin{tabular}[c]{@{}l@{}}416\\ 416\\ 416\end{tabular} & \begin{tabular}[c]{@{}l@{}}201\\ 204\\ 208\end{tabular} & \multicolumn{1}{l|}{\begin{tabular}[c]{@{}l@{}}2470.7\\ 618.2\\ 153.2\end{tabular}} & \begin{tabular}[c]{@{}l@{}}786.0\\ 199.3\\ 52.7\end{tabular} & 100\% & 47.9\% & \begin{tabular}[c]{@{}l@{}}3.1\\ 3.1\\ 2.9\end{tabular} & \begin{tabular}[c]{@{}l@{}}1.4\%\\ 1.8\%\\ 6.9\%\end{tabular} \\ \cline{2-21} 
 & 7\% & \begin{tabular}[c]{@{}l@{}}1\\ 4\\ 16\end{tabular} & 0.791 & \begin{tabular}[c]{@{}l@{}}434\\ 436\\ 448\end{tabular} & \begin{tabular}[c]{@{}l@{}}70\\ 72\\ 80\end{tabular} & \begin{tabular}[c]{@{}l@{}}2445.4\\ 606.2\\ 149.3\end{tabular} & \begin{tabular}[c]{@{}l@{}}251.8\\ 63.9\\ 18.0\end{tabular} & 51.9\% & 34.2\% & \begin{tabular}[c]{@{}l@{}}9.7\\ 9.5\\ 8.3\end{tabular} & \begin{tabular}[c]{@{}l@{}}2.7\%\\ 3.6\%\\ 12.7\%\end{tabular} & 0.766 & \begin{tabular}[c]{@{}l@{}}311\\ 312\\ 320\end{tabular} & \begin{tabular}[c]{@{}l@{}}129\\ 132\\ 144\end{tabular} & \multicolumn{1}{l|}{\begin{tabular}[c]{@{}l@{}}1822.2\\ 456.1\\ 116.2\end{tabular}} & \begin{tabular}[c]{@{}l@{}}503.2\\ 128.0\\ 36.4\end{tabular} & 56.0\% & 41.4\% & \begin{tabular}[c]{@{}l@{}}3.6\\ 3.6\\ 3.2\end{tabular} & \begin{tabular}[c]{@{}l@{}}2.2\%\\ 2.8\%\\ 10.0\%\end{tabular} \\ \cline{2-21} 
 & 8\% & \begin{tabular}[c]{@{}l@{}}1\\ 4\\ 16\end{tabular} & 0.782 & \begin{tabular}[c]{@{}l@{}}297\\ 300\\ 304\end{tabular} & \begin{tabular}[c]{@{}l@{}}11\\ 12\\ 16\end{tabular} & \begin{tabular}[c]{@{}l@{}}1632.8\\ 411.7\\ 102.4\end{tabular} & \begin{tabular}[c]{@{}l@{}}42.3\\ 10.1\\ 3.2\end{tabular} & 45.4\% & 30.4\% & \begin{tabular}[c]{@{}l@{}}38.6\\ 40.8\\ 32.0\end{tabular} & \begin{tabular}[c]{@{}l@{}}16.2\%\\ 22.7\%\\ 71.6\%\end{tabular} & 0.756 & \begin{tabular}[c]{@{}l@{}}201\\ 204\\ 208\end{tabular} & \begin{tabular}[c]{@{}l@{}}82\\ 84\\ 96\end{tabular} & \multicolumn{1}{l|}{\begin{tabular}[c]{@{}l@{}}1164.1\\ 294.8\\ 75.0\end{tabular}} & \begin{tabular}[c]{@{}l@{}}322.9\\ 83.1\\ 26.1\end{tabular} & 47.9\% & 39.0\% & \begin{tabular}[c]{@{}l@{}}3.6\\ 3.5\\ 2.9\end{tabular} & \begin{tabular}[c]{@{}l@{}}3.4\%\\ 4.4\%\\ 13.9\%\end{tabular} \\ \hline
\end{tabular}\\
  * \textit{thr\_acc}: accuracy corresponding to an accuracy drop rate $\alpha$.\hspace{.1in} 
  \textit{base}: baseline approach.\hspace{.1in} \textit{comp}: composability-based approach.\\
  \textit{speedup}: $Time_{base}/Time_{comp}$; overhead counted in $Time_{comp}$.\hspace{.1in}
  \textit{overhead}: block training time over the total time of \textit{comp}. 
\end{table*}

\begin{table}
  \centering
  \scriptsize
 \tabcolsep=0.04cm 
 \caption{\blue{Speedups  by composability-based pruning with different subspace sizes.}}
 \vspace{-.5\baselineskip}
 \label{tab:different_subspace_size}
\begin{tabular}{|l|l|l|l|l|l||l|l|l|}
\hline
\multirow{2}{*}{Dataset} & \multirow{2}{*}{alpha} & \multirow{2}{*}{\begin{tabular}[c]{@{}l@{}}subspace \\ size\end{tabular}} & \multicolumn{3}{l||}{ResNet-50} & \multicolumn{3}{l|}{Inception-V3} \\ \cline{4-9} 
 &  &  & \begin{tabular}[c]{@{}l@{}}base\\ time (h)\end{tabular} & \begin{tabular}[c]{@{}l@{}}comp\\ time (h)\end{tabular} & \begin{tabular}[c]{@{}l@{}}speedup\\ (X)\end{tabular} & \begin{tabular}[c]{@{}l@{}}base\\ time (h)\end{tabular} & \begin{tabular}[c]{@{}l@{}}comp\\ time (h)\end{tabular} & \begin{tabular}[c]{@{}l@{}}speedup\\ (X)\end{tabular} \\ \hline
\multirow{4}{*}{Flowers102} & \multirow{4}{*}{0\%} & 4 & 22.7 & 13.4 & 1.7 & 20.3 & 16.8 & 1.2 \\ \cline{3-9} 
 &  & 16 & 90.9 & 12.8 & 7.1 & 76.7 & 20.6 & 3.7 \\ \cline{3-9} 
 &  & 64 & 364.8 & 21 & 17.4 & 224.7 & 25.4 & 8.8 \\ \cline{3-9} 
 &  & 256 & 1460.7 & 13.5 & 108.2 & 809.4 & 40.7 & 19.9 \\ \hline\hline
\multirow{4}{*}{CUB200} & \multirow{4}{*}{3\%} & 4 & 22.8 & 11 & 2.1 & 23.6 & 26 & 0.9 \\ \cline{3-9} 
 &  & 16 & 93.8 & 11.4 & 8.2 & 83.5 & 30 & 2.8 \\ \cline{3-9} 
 &  & 64 & 369.6 & 15.5 & 23.8 & 292.5 & 29.2 & 10 \\ \cline{3-9} 
 &  & 256 & 1472.9 & 20.7 & 71.2 & 1128.9 & 18.1 & 62.4 \\ \hline
\end{tabular}
	\vspace{-.1in}
\end{table}

\begin{table}
  \centering
  \scriptsize
 \tabcolsep=0.04cm 
  \caption{Extra speedups brought by improved tuning block definitions.}
\vspace{-.5\baselineskip}
  \label{tab:extra_speedups}
  \begin{tabular}{|l|l||l|l|l||l|l|l|}
    \hline
    \multirow{3}{*}{Dataset} & \multirow{3}{*}{$\alpha$} & \multicolumn{3}{l||}{ResNet-50} & \multicolumn{3}{l|}{Inception-V3} \\ \cline{3-8} 
    &  & \multirow{2}{*}{thr\_acc} & \multicolumn{2}{l||}{extra speedup (X)} & \multirow{2}{*}{thr\_acc} & \multicolumn{2}{l|}{extra speedup (X)} \\ \cline{4-5} \cline{7-8} 
    &  &  & collection-1 & collection-2 &  & collection-1 & collection-2 \\ \hline
    \multirow{3}{*}{Flowers102} & 0\% & 0.973 & 1.05 & 0.98 & 0.968 & 1.12 & 1.14 \\ \cline{2-8} 
    & 1\% & 0.963 & 1.19 & 1.21 & 0.958 & 1.08 & 1.15 \\ \cline{2-8} 
    & 2\% & 0.953 & 1.06 & 1.14 & 0.949 & 1.15 & 1.23 \\ \hline
    \multirow{3}{*}{CUB200} & 3\% & 0.747 & 1.04 & 1.08 & 0.737 & 1.00 & 1.03 \\ \cline{2-8} 
    & 4\% & 0.739 & 1.04 & 1.20 & 0.729 & 1.08 & 1.09 \\ \cline{2-8} 
    & 5\% & 0.731 & 1.11 & 1.15 & 0.722 & 1.03 & 1.04 \\ \hline \hline
    \multicolumn{3}{|l|}{geometric mean } & 1.08 & 1.12 &  & 1.08 & 1.11 \\ \hline
  \end{tabular}\\
	\vspace*{-\baselineskip}
\end{table}

\paragraph{Extra Benefits from Tuning Blocks Identification}

Hierarchical tuning block identifier balances the overhead of training tuning blocks and the time savings they bring to the fine-tuning of pruned networks. 
Table~\ref{tab:extra_speedups} reports the extra speedups brought when it is used.   

For datasets Flowers102 and CUB200, we experiment with two types of collections of configurations with $N=8$. The first type, ``collection-1'', is a randomly sampled collection as mentioned earlier, and the second type, ``collection-2'', is attained by setting one pruning rate for a sequence of convolution modules, similar to the prior work~\cite{li2016pruning} to reduce module-wise meta-parameters. 
For each type, we repeat the experiments five times with a new collection created each time. 
Each tuning block identified from the first collection tends to contain only one convolution module due to the independence in choosing the pruning rate for each module. But the average number of tuning blocks is less than the total number of possible pruned convolution modules (41 versus 48 for ResNet-50 and 27 versus 33 for Inception-V3) because of the small collection size.  The "collection-2" setting has tuning blocks that contain a sequence of convolution modules as they are set to use one pruning rate.  

The extra speedups from the algorithm are substantial for both, but more on the "collection-2" setting for the opportunities that some larger popular tuning blocks have for benefiting the networks in that collection. Because some tuning blocks selected by the algorithm are a sequence of convolution modules that frequently appear in the collections, the total number of tuning blocks becomes smaller (e.g., 27 versus 23 on Inception-V3.) 

\section{Conclusions and Future Work}

By drawing on the recent framework CoCoPIE, this article has introduced the concept of {\em compression-compilation co-design} and how it is materialized into a software framework CoCoPIE for real-time AI on mobile devices. The results produced by the two core components, CoCo-Gen and CoCo-Tune, provide strong evidences for the promise of the co-design principle.  They indicate that it is possible to instill AI directly on existing commodity computing devices while offering even higher speeds and better energy efficiency than special AI accelerating hardware. The results open new opportunities for democratizing AI capability on end devices, while invalidating the common perception on the indispensability of special AI hardware for real-time AI on end devices. We believe that these results will prompt the industry to reexamine the directions and strategies on the pursue of mobile AI.


The promising progress opens up many potential directions for future development. We list two of them here.

The first is to expand the scope of the co-design based optimizations. So far, the principle of compression-compilation co-design has been focused on DNN models. Besides DNN, a real-world AI application often includes a lot of other parts, such as data collection, data preprocessing, the use of the DNN prediction in follow-up operations, and so on. Even though DNN may play an important role in the overall application, its optimizations may not be sufficient for the entire application to meet users' needs. So an important direction is on how to generalize the co-design principle into holistic optimizations to the entire AI-based applications. 

The second is to increase the applicability of the co-design based optimizations. This direction relates with privacy and security. As they are two important factors in many AI model constructions and deployments, how to integrate them into the co-design process is worth pursuing. For instance, typically model pruning requires access to both the models and the training dataset, but there are scenarios where datasets may not be accessible to the model optimizer due to either privacy policies or artificial boundaries among corporations. Effective ways to circumvent these roadblocks could expand the applicability of the optimizations. This direction also relates with the way that the optimization framework takes to deliver its service (e.g., standalone software versus cloud-based service). 

Following these directions, we envision the following roadmap of CoCoPIE development such that it can better serve real-world applications. First, we plan to develop it into a full-fledged DNN optimizing framework for mobile AI, including a versatile front end to support DNNs written in all the popular programming frameworks and an even broader coverage of DNN models and operations. Second, we plan to make CoCoPIE more flexible in delivering its service to fit the needs of various practical settings. Besides making it a standalone software framework, we envision a model of {\em DNN optimization as a service}, where, users may use CoCoPIE as a cloud-based service. It could not only lower the barrier for users to start adopting the service, but also avoid the hassles in setting up servers for DNN pruning. Moreover, based on our recent work~\cite{zheng2020priv}, we plan to explore {\em data-free DNN pruning}, which could prune a DNN with generated rather than original training data, making the technique applicable to situations where training data are hard to access. Finally, after meeting the needs of most DNNs, we plan to extend CoCoPIE into an application-level optimizing framework by providing a holistic treatment to the efficiency issues in the entire AI applications as well as the whole (typically heterogeneous) deployment system.

\paragraph{Final Words on CoCoPIE:} The authors are actively applying CoCoPIE to meet practical needs. CoCoPIE’s technology can immediately enable real-time deep learning on billions of existing mobile devices, thus generating tremendous commercial values.  To just name a few,  CoCoPIE may enable great user experiences for streaming applications, such as YouTube, TikTok, and Snap, even under low-bandwidth situations: These applications can stream low-resolution videos to user devices, and CoCoPIE can upscale the videos to high-definition in real time. Similarly, video communication applications such as Zoom, Skype, and Webex, can utilize CoCoPIE’s technology to deliver the best quality of service.  In addition, CoCoPIE unlocks real-time deep learning applications that have never been possible before, such as enabling a mobile phone camera to show live videos in an artistic style. The authors welcome business ideas, suggestions, or any comments (contact: info@cocopie.ai).  


\bibliography{wootzRef,all}

\begin{thebibliography}{10}

\bibitem{caffeSolver}
{Caffe Solver Prototxt}.
\newblock \url{https://github.com/BVLC/caffe/wiki/Solver-Prototxt}.

\bibitem{Ali-MNN}
Alibaba.
\newblock Mnn, 2019.

\bibitem{ashok2017n2n}
Anubhav Ashok, Nicholas Rhinehart, Fares Beainy, and Kris~M Kitani.
\newblock N2n learning: Network to network compression via policy gradient
  reinforcement learning.
\newblock {\em arXiv preprint arXiv:1709.06030}, 2017.

\bibitem{ba2014deep}
Jimmy Ba and Rich Caruana.
\newblock Do deep nets really need to be deep?
\newblock In {\em Advances in neural information processing systems}, pages
  2654--2662, 2014.

\bibitem{bergstra2012random}
James Bergstra and Yoshua Bengio.
\newblock Random search for hyper-parameter optimization.
\newblock {\em Journal of Machine Learning Research}, 13(Feb):281--305, 2012.

\bibitem{bucilua2006model}
Cristian Bucilua, Rich Caruana, and Alexandru Niculescu-Mizil.
\newblock Model compression.
\newblock In {\em Proceedings of the 12th ACM SIGKDD international conference
  on Knowledge discovery and data mining}, pages 535--541. ACM, 2006.

\bibitem{chen2018tvm}
Tianqi Chen, Thierry Moreau, Ziheng Jiang, Lianmin Zheng, Eddie Yan, Haichen
  Shen, Meghan Cowan, Leyuan Wang, Yuwei Hu, Luis Ceze, et~al.
\newblock {TVM}: An automated end-to-end optimizing compiler for deep learning.
\newblock In {\em 13th USENIX Symposium on Operating Systems Design and
  Implementation (OSDI 18)}, pages 578--594, 2018.

\bibitem{isscc_2016_chen_eyeriss}
{Chen, Yu-Hsin and Krishna, Tushar and Emer, Joel and Sze, Vivienne}.
\newblock {Eyeriss: An Energy-Efficient Reconfigurable Accelerator for Deep
  Convolutional Neural Networks}.
\newblock In {\em {IEEE International Solid-State Circuits Conference, ISSCC
  2016, Digest of Technical Papers}}, pages {262--263}, {2016}.

\bibitem{courbariaux2015binaryconnect}
Matthieu Courbariaux, Yoshua Bengio, and Jean-Pierre David.
\newblock Binaryconnect: Training deep neural networks with binary weights
  during propagations.
\newblock In {\em Advances in neural information processing systems}, pages
  3123--3131, 2015.

\bibitem{dai2017nest}
Xiaoliang Dai, Hongxu Yin, and Niraj~K Jha.
\newblock Nest: a neural network synthesis tool based on a grow-and-prune
  paradigm.
\newblock {\em arXiv preprint arXiv:1711.02017}, 2017.

\bibitem{dong2014learning}
Chao Dong, Chen~Change Loy, Kaiming He, and Xiaoou Tang.
\newblock Learning a deep convolutional network for image super-resolution.
\newblock In {\em European conference on computer vision}, pages 184--199.
  Springer, 2014.

\bibitem{fu2017look}
Jianlong Fu, Heliang Zheng, and Tao Mei.
\newblock Look closer to see better: Recurrent attention convolutional neural
  network for fine-grained image recognition.
\newblock In {\em Conf. on Computer Vision and Pattern Recognition}, 2017.

\bibitem{gatys2016image}
Leon~A Gatys, Alexander~S Ecker, and Matthias Bethge.
\newblock Image style transfer using convolutional neural networks.
\newblock In {\em Proceedings of the IEEE conference on computer vision and
  pattern recognition}, pages 2414--2423, 2016.

\bibitem{TensorFlow-Lite}
Google.
\newblock Tensorflow lite, 2019.

\bibitem{googletpu}
{Google Cloud TPU}.
\newblock Google cloud tpu.
\newblock \url{https://cloud.google.com/tpu/}, 2017.

\bibitem{Guan+:PLDI2019}
Hui Guan, Xipeng Shen, and Seung-Hwan Lim.
\newblock Wootz: A compiler-based framework for fast cnn pruning via
  composability.
\newblock In {\em Proceedings of the Programming Language Design and
  Implementation (PLDI)}, 2019.

\bibitem{guo2016dynamic}
Yiwen Guo, Anbang Yao, and Yurong Chen.
\newblock Dynamic network surgery for efficient dnns.
\newblock In {\em Advances in neural information processing systems}, pages
  1379--1387, 2016.

\bibitem{han2017ese}
Song Han, Junlong Kang, Huizi Mao, Yiming Hu, Xin Li, Yubin Li, Dongliang Xie,
  Hong Luo, Song Yao, Yu~Wang, Huazhong Yang, and William~J. Dally.
\newblock Ese: Efficient speech recognition engine with sparse lstm on fpga.
\newblock In {\em FPGA}, pages 75--84, 2017.

\bibitem{han2015learning}
Song Han, Jeff Pool, John Tran, and William Dally.
\newblock Learning both weights and connections for efficient neural network.
\newblock In {\em Advances in neural information processing systems}, pages
  1135--1143, 2015.

\bibitem{he2016deep}
Kaiming He, Xiangyu Zhang, Shaoqing Ren, and Jian Sun.
\newblock Deep residual learning for image recognition.
\newblock In {\em Proceedings of the IEEE conference on computer vision and
  pattern recognition}, pages 770--778, 2016.

\bibitem{he2018adc}
Yihui He and Song Han.
\newblock Adc: Automated deep compression and acceleration with reinforcement
  learning.
\newblock {\em arXiv preprint arXiv:1802.03494}, 2018.

\bibitem{he2017channel}
Yihui He, Xiangyu Zhang, and Jian Sun.
\newblock Channel pruning for accelerating very deep neural networks.
\newblock In {\em Computer Vision (ICCV), 2017 IEEE International Conference
  on}, pages 1398--1406. IEEE, 2017.

\bibitem{hinton2015distilling}
Geoffrey Hinton, Oriol Vinyals, and Jeff Dean.
\newblock Distilling the knowledge in a neural network.
\newblock {\em arXiv preprint arXiv:1503.02531}, 2015.

\bibitem{hoos2011automated}
Holger~H Hoos.
\newblock Automated algorithm configuration and parameter tuning.
\newblock In {\em Autonomous search}, pages 37--71. Springer, 2011.

\bibitem{howard2017mobilenets}
Andrew~G Howard, Menglong Zhu, Bo~Chen, Dmitry Kalenichenko, Weijun Wang,
  Tobias Weyand, Marco Andreetto, and Hartwig Adam.
\newblock Mobilenets: Efficient convolutional neural networks for mobile vision
  applications.
\newblock {\em arXiv preprint arXiv:1704.04861}, 2017.

\bibitem{hu2016network}
Hengyuan Hu, Rui Peng, Yu-Wing Tai, and Chi-Keung Tang.
\newblock Network trimming: A data-driven neuron pruning approach towards
  efficient deep architectures.
\newblock {\em arXiv preprint arXiv:1607.03250}, 2016.

\bibitem{hubara2016binarized}
Itay Hubara, Matthieu Courbariaux, Daniel Soudry, Ran El-Yaniv, and Yoshua
  Bengio.
\newblock Binarized neural networks.
\newblock In {\em Advances in neural information processing systems}, pages
  4107--4115, 2016.

\bibitem{2925974}
Satoshi Iizuka, Edgar Simo-Serra, and Hiroshi Ishikawa.
\newblock Let there be color! joint end-to-end learning of global and local
  image priors for automatic image colorization with simultaneous
  classification.
\newblock {\em ACM Trans. Graph.}, 35(4), July 2016.

\bibitem{jia2014caffe}
Yangqing Jia, Evan Shelhamer, Jeff Donahue, Sergey Karayev, Jonathan Long, Ross
  Girshick, Sergio Guadarrama, and Trevor Darrell.
\newblock Caffe: Convolutional architecture for fast feature embedding.
\newblock In {\em Proceedings of the 22nd ACM international conference on
  Multimedia}, pages 675--678. ACM, 2014.

\bibitem{khosla2011novel}
Aditya Khosla, Nityananda Jayadevaprakash, Bangpeng Yao, and Fei-Fei Li.
\newblock Novel dataset for fine-grained image categorization: Stanford dogs.
\newblock In {\em Proc. CVPR Workshop on Fine-Grained Visual Categorization
  (FGVC)}, volume~2, page~1, 2011.

\bibitem{krause2016unreasonable}
Jonathan Krause, Benjamin Sapp, Andrew Howard, Howard Zhou, Alexander Toshev,
  Tom Duerig, James Philbin, and Li~Fei-Fei.
\newblock The unreasonable effectiveness of noisy data for fine-grained
  recognition.
\newblock In {\em European Conference on Computer Vision}, pages 301--320.
  Springer, 2016.

\bibitem{krause20133d}
Jonathan Krause, Michael Stark, Jia Deng, and Li~Fei-Fei.
\newblock 3d object representations for fine-grained categorization.
\newblock In {\em Computer Vision Workshops (ICCVW), 2013 IEEE International
  Conference on}, pages 554--561. IEEE, 2013.

\bibitem{lavin2016fast}
Andrew Lavin and Scott Gray.
\newblock Fast algorithms for convolutional neural networks.
\newblock In {\em Proceedings of the IEEE Conference on Computer Vision and
  Pattern Recognition}, pages 4013--4021, 2016.

\bibitem{lebedev2016fast}
Vadim Lebedev and Victor Lempitsky.
\newblock Fast convnets using group-wise brain damage.
\newblock In {\em Proceedings of the IEEE Conference on Computer Vision and
  Pattern Recognition}, pages 2554--2564, 2016.

\bibitem{leng2017extremely}
Cong Leng, Hao Li, Shenghuo Zhu, and Rong Jin.
\newblock Extremely low bit neural network: Squeeze the last bit out with admm.
\newblock {\em arXiv preprint arXiv:1707.09870}, 2017.

\bibitem{li2016pruning}
Hao Li, Asim Kadav, Igor Durdanovic, Hanan Samet, and Hans~Peter Graf.
\newblock Pruning filters for efficient convnets.
\newblock {\em arXiv preprint arXiv:1608.08710}, 2016.

\bibitem{li2017pruning}
Hao Li, Asim Kadav, Igor Durdanovic, Hanan Samet, and Hans~Peter Graf.
\newblock Pruning filters for efficient convnets.
\newblock In {\em International Conference on Learning Representations (ICLR)},
  2017.

\bibitem{lin2016fixed}
Darryl Lin, Sachin Talathi, and Sreekanth Annapureddy.
\newblock Fixed point quantization of deep convolutional networks.
\newblock In {\em International Conference on Machine Learning}, pages
  2849--2858, 2016.

\bibitem{lin2014microsoft}
Tsung-Yi Lin, Michael Maire, Serge Belongie, James Hays, Pietro Perona, Deva
  Ramanan, Piotr Doll{\'a}r, and C~Lawrence Zitnick.
\newblock Microsoft coco: Common objects in context.
\newblock In {\em European conference on computer vision}, pages 740--755.
  Springer, 2014.

\bibitem{luo2017thinet}
Jian-Hao Luo, Jianxin Wu, and Weiyao Lin.
\newblock Thinet: A filter level pruning method for deep neural network
  compression.
\newblock {\em arXiv preprint arXiv:1707.06342}, 2017.

\bibitem{ma2019pconv}
Xiaolong Ma, Fu-Ming Guo, Wei Niu, Xue Lin, Jian Tang, Kaisheng Ma, Bin Ren,
  and Yanzhi Wang.
\newblock Pconv: The missing but desirable sparsity in dnn weight pruning for
  real-time execution on mobile devices.
\newblock {\em AAAI}, 2020.

\bibitem{mao2017exploring}
Huizi Mao, Song Han, Jeff Pool, Wenshuo Li, Xingyu Liu, Yu~Wang, and William~J
  Dally.
\newblock Exploring the regularity of sparse structure in convolutional neural
  networks.
\newblock {\em arXiv preprint arXiv:1705.08922}, 2017.

\bibitem{molchanov2016pruning}
Pavlo Molchanov, Stephen Tyree, Tero Karras, Timo Aila, and Jan Kautz.
\newblock Pruning convolutional neural networks for resource efficient transfer
  learning.
\newblock {\em arXiv preprint arXiv:1611.06440}, 2016.

\bibitem{nevill1997identifying}
Craig~G. Nevill-Manning and Ian~H. Witten.
\newblock {Identifying hierarchical structure in sequences: A linear-time
  algorithm}.
\newblock {\em J. Artif. Intell. Res.(JAIR)}, 7:67--82, 1997.

\bibitem{nilsback2008automated}
Maria-Elena Nilsback and Andrew Zisserman.
\newblock Automated flower classification over a large number of classes.
\newblock In {\em Computer Vision, Graphics \& Image Processing, 2008.
  ICVGIP'08. Sixth Indian Conference on}, pages 722--729. IEEE, 2008.

\bibitem{niu2020patdnn}
Wei Niu, Xiaolong Ma, Sheng Lin, Shihao Wang, Xuehai Qian, Xue Lin, Yanzhi
  Wang, and Bin Ren.
\newblock Patdnn: Achieving real-time dnn execution on mobile devices with
  pattern-based weight pruning.
\newblock {\em ASPLOS}, 2020.

\bibitem{park2017weighted}
Eunhyeok Park, Junwhan Ahn, and Sungjoo Yoo.
\newblock Weighted-entropy-based quantization for deep neural networks.
\newblock In {\em Proceedings of the IEEE Conference on Computer Vision and
  Pattern Recognition}, pages 7197--7205, 2017.

\bibitem{rastegari2016xnor}
Mohammad Rastegari, Vicente Ordonez, Joseph Redmon, and Ali Farhadi.
\newblock Xnor-net: Imagenet classification using binary convolutional neural
  networks.
\newblock In {\em European Conference on Computer Vision}, pages 525--542.
  Springer, 2016.

\bibitem{russakovsky2015imagenet}
Olga Russakovsky, Jia Deng, Hao Su, Jonathan Krause, Sanjeev Satheesh, Sean Ma,
  Zhiheng Huang, Andrej Karpathy, Aditya Khosla, Michael Bernstein,
  Alexander~C. Berg, and Li~Fei-Fei.
\newblock Imagenet large scale visual recognition challenge.
\newblock {\em International Journal of Computer Vision (IJCV)},
  115(3):211--252, 2015.

\bibitem{slim}
N.~Silberman and S.~Guadarrama.
\newblock Tensorflow-slim image classification model library.
\newblock \url{https://github.com/tensorflow/models/tree/master/research/slim},
  2016.

\bibitem{szegedy2015going}
Christian Szegedy, Wei Liu, Yangqing Jia, Pierre Sermanet, Scott Reed, Dragomir
  Anguelov, Dumitru Erhan, Vincent Vanhoucke, and Andrew Rabinovich.
\newblock Going deeper with convolutions.
\newblock In {\em Proceedings of the IEEE conference on computer vision and
  pattern recognition}, pages 1--9, 2015.

\bibitem{timofte2017ntire}
Radu Timofte, Eirikur Agustsson, Luc Van~Gool, Ming-Hsuan Yang, and Lei Zhang.
\newblock Ntire 2017 challenge on single image super-resolution: Methods and
  results.
\newblock In {\em Proceedings of the IEEE conference on computer vision and
  pattern recognition workshops}, pages 114--125, 2017.

\bibitem{welinder2010caltech}
Peter Welinder, Steve Branson, Takeshi Mita, Catherine Wah, Florian Schroff,
  Serge Belongie, and Pietro Perona.
\newblock Caltech-ucsd birds 200.
\newblock 2010.

\bibitem{wen2016learning}
Wei Wen, Chunpeng Wu, Yandan Wang, Yiran Chen, and Hai Li.
\newblock Learning structured sparsity in deep neural networks.
\newblock In {\em Advances in neural information processing systems}, pages
  2074--2082, 2016.

\bibitem{winograd1980arithmetic}
Shmuel Winograd.
\newblock {\em Arithmetic complexity of computations}, volume~33.
\newblock Siam, 1980.

\bibitem{wu2016quantized}
Jiaxiang Wu, Cong Leng, Yuhang Wang, Qinghao Hu, and Jian Cheng.
\newblock Quantized convolutional neural networks for mobile devices.
\newblock In {\em Proceedings of the IEEE Conference on Computer Vision and
  Pattern Recognition}, pages 4820--4828, 2016.

\bibitem{yamins2016using}
Daniel~LK Yamins and James~J DiCarlo.
\newblock Using goal-driven deep learning models to understand sensory cortex.
\newblock {\em Nature neuroscience}, 19(3):356, 2016.

\bibitem{yamins2014performance}
Daniel~LK Yamins, Ha~Hong, Charles~F Cadieu, Ethan~A Solomon, Darren Seibert,
  and James~J DiCarlo.
\newblock Performance-optimized hierarchical models predict neural responses in
  higher visual cortex.
\newblock {\em Proceedings of the National Academy of Sciences},
  111(23):8619--8624, 2014.

\bibitem{yu2018wide}
Jiahui Yu, Yuchen Fan, Jianchao Yang, Ning Xu, Zhaowen Wang, Xinchao Wang, and
  Thomas Huang.
\newblock Wide activation for efficient and accurate image super-resolution.
\newblock {\em arXiv preprint arXiv:1808.08718}, 2018.

\bibitem{zheng2020priv}
Zheng Zhan, Yifan Gong, Zhengang Li, Pu~Zhao, Xiaolong Ma, Wei Niu, Xiaolin Xu,
  Bin Ren, Yanzhi Wang, and Xue Lin.
\newblock Priv: A privacy-preserving deep neural network model compression
  framework.
\newblock {\em arXiv preprint}, 2020.

\bibitem{zhang2017multistyle}
Hang Zhang and Kristin Dana.
\newblock Multi-style generative network for real-time transfer.
\newblock {\em arXiv preprint arXiv:1703.06953}, 2017.

\bibitem{zhao2017diversified}
Bo~Zhao, Xiao Wu, Jiashi Feng, Qiang Peng, and Shuicheng Yan.
\newblock Diversified visual attention networks for fine-grained object
  classification.
\newblock {\em IEEE Transactions on Multimedia}, 19(6):1245--1256, 2017.

\bibitem{zhou2017incremental}
Aojun Zhou, Anbang Yao, Yiwen Guo, Lin Xu, and Yurong Chen.
\newblock Incremental network quantization: Towards lossless cnns with
  low-precision weights.
\newblock In {\em International Conference on Learning Representations (ICLR)},
  2017.

\bibitem{zhou2014learning}
Bolei Zhou, Agata Lapedriza, Jianxiong Xiao, Antonio Torralba, and Aude Oliva.
\newblock Learning deep features for scene recognition using places database.
\newblock In {\em Advances in neural information processing systems}, pages
  487--495, 2014.

\end{thebibliography}

\end{document}